\documentclass[10pt,a4,oneside,onecolumn]{article}

\usepackage{hyperref}
\makeatletter
\def\cl@chapter{}
\makeatother
\usepackage[capitalise,nameinlink]{cleveref} % in-document referencing

\usepackage{setspace}
\usepackage{geometry}
\usepackage{multicol}
\usepackage{setspace}
\usepackage[utf8]{inputenc}
\usepackage[T1]{fontenc}
\usepackage[]{microtype}
\usepackage[]{natbib}

\usepackage{color,graphicx}
\usepackage[usenames,dvipsnames]{xcolor}

\usepackage{mathtools} % redefine operators
\usepackage{amssymb}   % needed for mathbb
\usepackage{amsfonts}  % needed for mathbb
\usepackage{mathrsfs}  % needed for mathscr
\usepackage[group-separator={,}, detect-weight=true, detect-family=true]{siunitx}   % replace sistyle, siunits, units
\usepackage{tabu,booktabs,tabularx} % tables
\usepackage{csquotes} % quotes
\usepackage{textcase}
\usepackage[explicit,pagestyles]{titlesec}
\usepackage{parskip} % no parident
\usepackage{xparse}
\usepackage{subcaption}
\usepackage{xfrac}

\usepackage[sc,osf]{mathpazo}   % With old-style figures and real smallcaps.
\usepackage[euler-digits,small]{eulervm}
\usepackage{xspace}

\usepackage{pifont}%
\newcommand{\cmark}{\ding{51}}%
\newcommand{\xmark}{\ding{55}}%

\usepackage[thmmarks]{ntheorem}
\crefname{Theorem}{Theorem}{Theorems}
\crefname{Corollary}{Corollary}{Corollaries}
\crefname{Lemma}{Lemma}{Lemmas}
\crefname{Definition}{Definition}{Definitions}
\crefname{Proposition}{Proposition}{Propositions}
\crefname{Problem}{Problem}{Problems}
\crefname{Proof}{Proof}{Proofs}
\crefname{Problem}{Problem}{Problem}

\makeatletter
\newtheoremstyle{msplain}%
{\item[\hskip\labelsep \theorem@headerfont ##1\ ##2\theorem@separator]}%
{\item[\hskip\labelsep \theorem@headerfont ##1\ ##2\ \normalfont{(\textit{##3})}\theorem@separator]}

\newtheoremstyle{msnonumberplain}%
{\item[\hskip\labelsep \theorem@headerfont ##1\theorem@separator]}%
{\item[\hskip\labelsep \theorem@headerfont ##3\theorem@separator]}%
\makeatother

\theoremnumbering{arabic}
\theoremstyle{msplain}
\theoremsymbol{\guillemotleft}
\theorembodyfont{\normalfont}
\theoremheaderfont{\normalfont\sffamily\bfseries}
\theoremseparator{:}
\theorempreskip{.5em} 
\theorempostskip{0em} 
\newtheorem{Theorem}{Theorem}%[section]
\newtheorem{Proposition}{Proposition}
\newtheorem{Lemma}{Lemma}

\newtheorem{Definition}{Definition}

\theoremstyle{msnonumberplain}
\theoremheaderfont{\normalfont\sffamily\bfseries}
\theorembodyfont{\normalfont}
\theoremsymbol{\ensuremath{_\Box}}
\newtheorem{Proof}{Proof}
\qedsymbol{\ensuremath{_\Box}}

\DeclareRobustCommand{\spacedallcaps}[1]{\sffamily{\MakeTextUppercase{#1}}}%
\DeclareRobustCommand{\spacedlowsmallcaps}[1]{\sffamily{\scshape\MakeTextLowercase{#1}}}%

\titleformat{\section}{\raggedright\normalfont\Large\sffamily}{\thesection}{.5em}{{#1}}
\titleformat{\subsection}{\raggedright\normalfont\large\sffamily}{\thesubsection}{1em}{\spacedallcaps{#1}}
\titleformat{\subsubsection}{\normalfont\sffamily}{\thesubsubsection}{1em}{\spacedallcaps{#1}}
\titleformat{\paragraph}{\normalfont\normalsize\sffamily}{\textsc{\MakeTextLowercase{\theparagraph}}}{0pt}{\spacedlowsmallcaps{#1}} 

\graphicspath{{./fig/}}

\newcommand{\maxf}[1]{{\underline{#1}}}

\renewcommand{\Pr}     {\ensuremath{P}}                 % Pr
\newcommand{\PP}     {\ensuremath{\mathbb{P}}}                 % P
\newcommand{\Real}   {\ensuremath{\mathbb{R}}}                 % Real numbers R
\newcommand{\Nat}   {\ensuremath{\mathbb{N}}}                 % Real numbers R
\newcommand{\euler}  {\ensuremath{\mathrm{e}}}                 % e
\newcommand{\im}     {\ensuremath{\mathrm{i}}}                 % i
\DeclarePairedDelimiterX{\norm}[1]{\lVert}{\rVert}{#1}         % norm
\DeclarePairedDelimiterX\innerprod[2]{\langle}{\rangle}{#1,#2} % inner product
\DeclareMathOperator{\ExpOp}{\mathbb{E}}                       % expectation 
\let\PM\PP
\DeclarePairedDelimiter\ip{\langle}{\rangle}
\newcommand*{\dd}{\mathop{\mathrm{d}\!}}
\newcommand{\mun}{\ensuremath{\mu[\PM_n]}}
\newcommand{\muP}{\ensuremath{\mu[\PM]}}
\newcommand{\define}{\emph}
\newcommand{\CN}{\smallcaps{C}_{\smallcaps{N}}}	% class of normal instances
\newcommand{\CA}{\smallcaps{C}_{\smallcaps{A}}}	% class of anomaly instances
\newcommand{\BigO}{\ensuremath{\mathcal{O}}}

\SetCiteCommand{\citep}

\renewcommand{\mkbegdispquote}[2]{\openautoquote}

\MakeOuterQuote{"}

\newcommand{\autocite}{\citep}

\definecolor{webgreen}           {rgb}{0   ,.5   ,0    }
\definecolor{webbrown}           {rgb}{.6  ,0    ,0    }
\hypersetup{%
	colorlinks=true, 
	urlcolor=black, linkcolor=black, citecolor=webgreen
}

\usepackage{textcase}
\newcommand{\smallcaps}[1]{\textsc{\MakeTextLowercase{#1}}}
\newcommand{\acronym}[1]{\smallcaps{#1}}
\DeclareRobustCommand{\EXPoSE}{\acronym{EXPOSE}\xspace}
\DeclareRobustCommand{\RKHS}{\acronym{RKHS}\xspace}

\newcommand{\hairsp}{\hspace{1pt}}% hair space
\newcommand{\ie}{\textit{i.\hairsp{}e.}\xspace}
\newcommand{\eg}{\textit{e.\hairsp{}g.}\xspace}
\providecommand\given{}

\DeclarePairedDelimiterX\SetDefX[1]{\lbrace}{\rbrace}{
	\renewcommand\given{  \nonscript\:
		\delimsize\vert
		\nonscript\:
		\mathopen{}
		\allowbreak}
	#1
}
\newcommand\SetDef{\SetDefX}

\DeclarePairedDelimiterX\conditionX[1]{}{}{
	\renewcommand\given{  \nonscript\:
		\delimsize\vert
		\nonscript\:
		\mathopen{}
		\allowbreak}
	#1
}
\newcommand\condition{\conditionX}

\usepackage{tcolorbox}
\tcbset{shield externalize}
\DeclareRobustCommand{\EXPoSE}{EXPoSE\xspace}
\newcommand\XX{\mathcal{X}}
\newcommand\XXX{\mathscr{X}}
\newcommand\HH{\mathcal{H}}

\begin{document}
	
	\title{\normalfont\spacedallcaps{Expected Similarity Estimation for Large-Scale Batch and Streaming Anomaly Detection}}
	\author{\spacedlowsmallcaps{Markus Schneider\textsuperscript{$\alpha,\beta$}, Wolfgang Ertel\textsuperscript{$\beta$}}\\
		\spacedlowsmallcaps{\& Fabio Ramos\textsuperscript{$\gamma$}}}
	\date{}
	
\maketitle
	
\begin{center}
	\begin{tcolorbox}[colback=white,width=\textwidth,colframe=white]
		\section*{Abstract}
		\small
		We present a novel algorithm for anomaly detection on very large datasets and data streams. The method, named \emph{EXPected Similarity Estimation} (\EXPoSE), is kernel-based and able to efficiently compute the similarity between new data points and the distribution of regular data. The estimator is formulated as an inner product with a reproducing kernel Hilbert space embedding and makes no assumption about the type or shape of the underlying data distribution. 
		We show that offline (batch) learning with \EXPoSE can be done in \emph{linear} time and online (incremental) learning takes \emph{constant} time per instance and model update. Furthermore, \EXPoSE can make predictions in \emph{constant} time, while it requires only \emph{constant} memory.
		In addition, we propose different methodologies for concept drift adaptation on evolving data streams. 
		On several real datasets we demonstrate that our approach can compete with state of the art algorithms for anomaly detection while being an order of magnitude faster than most other approaches.
	\end{tcolorbox}
\end{center}
	
	{\let\thefootnote\relax\footnotetext{\raggedright\textsuperscript{$\alpha$} \textit{Institute of Neural Information Processing,
			University of Ulm, Germany}}}
	{\let\thefootnote\relax\footnotetext{\raggedright\textsuperscript{$\beta$} \textit{Institute for Artificial Intelligence,
				Ravensburg-Weingarten University of Applied Sciences, Germany}}}
	{\let\thefootnote\relax\footnotetext{\raggedright\textsuperscript{$\gamma$} \textit{School of Information Technologies, The University of Sydney, Australia}}}
	{\let\thefootnote\relax\footnotetext{\raggedright\textit{This is an extended and revised version of a preliminary conference report that was presented in the International Joint Conference on Neural Networks 2015 \autocite{Schneider2015a}.}}}
	{\let\thefootnote\relax\footnotetext{\raggedright\textit{The final publication is available at Springer via \normalfont\url{http://dx.doi.org/10.1007/s10994-016-5567-7}}}}	
	
\section{Introduction}
What is an anomaly? An \emph{anomaly} is an element whose properties differ from the majority of other elements under consideration which are called the \emph{normal} data. \textcquote{Chandola2009}{\emph{Anomaly detection} refers to the problem of finding patterns in data that do not conform to expected behavior. These non-conforming patterns are often referred to as \emph{anomalies} \textelp{}}.

Typical applications of anomaly detection are network intrusion detection, credit card fraud detection, medical diagnosis and failure detection in industrial environments.
For example, systems which detect \emph{unusual} network behavior can be used to complement or replace traditional intrusion detection methods which are based on experts' knowledge in order to defeat the increasing number of attacks on computer based networks \autocite{Kumar2005}. Credit card transactions which differ significantly from the usual shopping behavior of the card owner can indicate that the credit card was stolen or a compromise of data associated with the account occurred \autocite{Aleskerov1997}.
The diagnosis of radiographs can be supported by automated systems to detect breast cancers in  mammographic image analysis \autocite{Spence2001}.
Unplanned downtime of production lines caused by failing components is a serious concern in many industrial environments. Here anomaly detection can be used to detect unusual sensor information to predict possible faults and enabling condition-based maintenance \autocite{zhang2011probabilistic}.
Novelty detection can be used to detect new interesting or unusual galaxies in astronomical data such as the Sloan Digital Sky Survey \autocite{Xiong2011}.

Obtaining labeled training data for all types of anomalies is often too expensive. Imagine the labeling has to be done by a human expert or is obtained through costly experiments \autocite{Hodge2004}. In some applications anomalies are also very rare as in air traffic safety or space missions. Hence, the problem of anomaly detection is typically unsupervised, however it is implicitly assumed that the dataset contains only very few anomalies. This assumption is reasonable since it is quite often possible to collect large amounts of data for the normal state of a system as, for example usual credit card transactions or network traffic of a system not under attack.

The computational complexity and memory requirements of classical algorithms become the limiting factor when applied to large-scale datasets as they occur nowadays. To solve this problem we propose a new anomaly detection algorithm called \emph{EXPected Similarity Estimation} (\EXPoSE).
As explained later in detail, the \EXPoSE anomaly detection classifier
\begin{align*}
\eta(z) = \ip{ \phi(z), \mu[\PM]}
\end{align*}
calculates a score (the likelihood of $z$ belonging to the class of normal data) using the inner product between a feature map $\phi$ and the kernel mean map $\mu[\PM]$ of the distribution of normal data $\PM$. We will show that this inner product can be evaluated in \emph{constant} time, while $\mu[\PM]$ can be estimated in \emph{linear} time, has \emph{constant} memory consumption and is designed to solve \emph{very} large-scale anomaly detection problems.

    \begin{figure}[tb]
    	\centering
    	\includegraphics{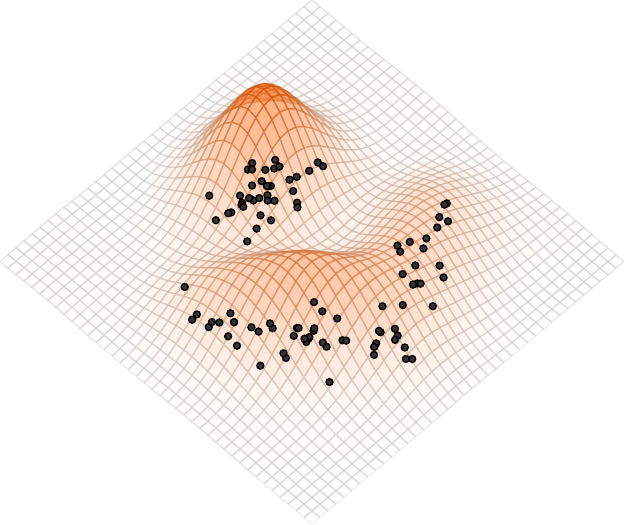}
    	\caption{Sketch of the \EXPoSE scores $\eta(z)$ in $\Real^2$, given some samples (black dots).}
    	\label{fig:expose_synthetic_contour}
    \end{figure}

Moreover, we will see that the proposed \EXPoSE classifier can be learned incrementally making it applicable to \emph{online} and \emph{streaming} anomaly detection problems. 
Learning on data streams directly is unavoidable in many applications such as network traffic monitoring, video surveillance and document feeds as data arrives continuously in fast streams with a volume too large or impractical to store.

Only a few anomaly detection algorithms can be applied to large-scale problems and even less are applicable to streaming data. The proposed \EXPoSE anomaly detector fills this gap.

\paragraph{Our main contributions are:}

\begin{itemize}
	\item We present an efficient anomaly detection algorithm, called \emph{EXPected Similarity Estimation} (\EXPoSE), with $\mathcal{O}(n)$ training time, $\mathcal{O}(1)$ prediction time and only $\mathcal{O}(1)$ memory requirements with respect to the dataset size $n$.
	\item We show that \EXPoSE is especially suitable for parallel and distributed processing which makes it scalable to very large problems.
	\item We demonstrate how \EXPoSE can be applied to online and streaming anomaly detection, while requiring only $\mathcal{O}(1)$ time for a model update, $\mathcal{O}(1)$ time per prediction and $\mathcal{O}(1)$ memory.
	\item We introduce two different approaches which allow \EXPoSE to be efficiently used with the most common techniques for concept drift adaptation.
	\item We evaluate \EXPoSE on several real datasets, including surveillance, image data and network intrusion detection.	
\end{itemize}

This paper is organised as follows: We first provide a formal problem description including a definition of batch and streaming anomaly detection. \Cref{sec:related_work} provides an overview of related work and a comparison of these techniques. \Cref{sec:expose} introduces the \EXPoSE anomaly detection algorithm along with the necessary theoretical framework. Subsequently we show in \cref{sec:oexpose} how \EXPoSE can be applied to streaming anomaly detection problems. The key to \EXPoSE s computational performance is subject to \cref{sec:featuremaps}. In \cref{experiments} we empirically compare \EXPoSE with several state of the art anomaly detectors.

\section{Problem Definition}

Even though there is a vast amount of literature on anomaly detection, there is no unique definition of what anomalies are and what exactly anomaly detection is. In this section we will state the problem of anomaly detection in batch and streaming applications.

\begin{Definition}[Input Space]
	The \define{input space} for an observation $X$ is a measurable space\footnote{A \define{measurable space} is a tuple $(\mathcal{X},\mathscr{X})$, where $\mathcal{X}$ is a nonempty set and $\mathscr{X}$ is a $\sigma$-algebra of its subsets. We refer the reader unfamiliar with this topic to \citet{kallenberg2006foundations} for an overview.} $(\mathcal{X},\mathscr{X})$ containing all values that $X$ might take. We denote the realization after measurement of the random variable $X$ with $X=x$.
\end{Definition}

We make no assumptions about the nature of the input space $\mathcal{X}$ which can consist of simple numerical vectors, but also can contain images, video data or trajectories of vehicles and people. We assume that there is a true (but unknown) distribution $\PP_X\colon \mathscr{X} \to [0,1]$ of the data.

\begin{figure}[tb]
	\centering
	\includegraphics{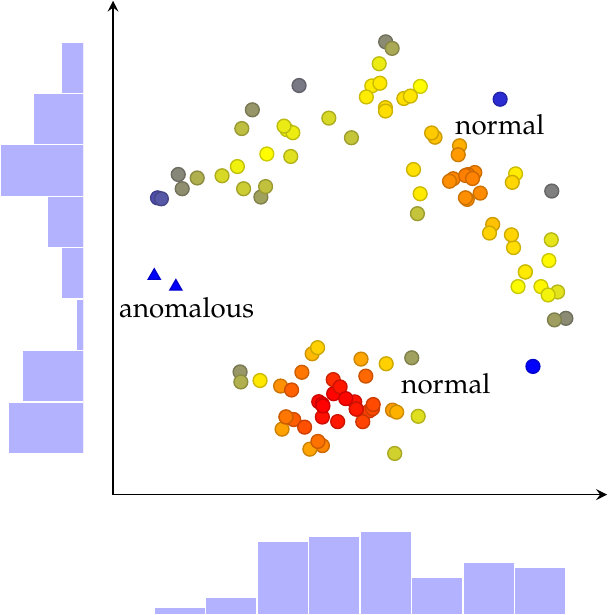}
	\caption[Normal distribution and anomalies]{Example of two instances (triangle) which are different from the distribution of normal data (circle) along with histograms of the marginal distributions.}
	\label{fig:anomaly_example}
\end{figure}

\begin{Definition}[Output/Label Space]
In anomaly detection an observation $X = x$ can belong to the class of normal data $\CN$ or can be an anomaly $\CA$. This is called \emph{label} of the observation and denoted by the random variable $Y$. The collection of all labels is given by the measurable space $(\mathcal{Y},\mathscr{Y})$ called \define{label space} or \define{output space} (\cref{fig:anomaly_example}).
\end{Definition}

The distribution of the observation $x\in\mathcal{X}$ is stochastic and depends on the label $Y$ and hence is distributed according to $\PP_{\condition{X\given Y}}$.

\begin{Definition}[Prediction/Decision Space]
	Based on the outcome $X=x$ of an observation, the objective of an anomaly detection algorithm is to make a prediction $\vartheta \in \mathcal{Q}$, where the measurable space $(\mathcal{Q},\mathscr{Q})$ is called the \define{prediction space} or sometimes \define{decision space}.
\end{Definition}

The prediction space $\mathcal{Q}$ is not necessarily equal to label space $\mathcal{Y}$. Especially in anomaly detection and classification many algorithms calculate a probability or a score for a label. Such a score is called \define{anomaly score} if it quantifies the likelihood of $x$ belonging to $\CA$ and \define{normal score} if it determines the degree of certainty to which $x$ belongs to $\CN$.

Scoring based algorithms are more flexible than techniques which assign hard class labels since anomalies can be ranked and prioritized according their score or a domain specific discrimination threshold can be applied to separate anomalies from normal data. 
For example we can define a mapping $\tau\colon(\mathcal{Q},\mathscr{Q})\to(\mathcal{Y},\mathscr{Y})$ as
\begin{align*}
\tau_\theta(\vartheta) = \begin{cases} 
\CN & \mbox{if } \vartheta > \theta \\ 
\CA & \mbox{else}
\end{cases}
\end{align*}
based on the threshold $\theta$.
Such a domain specific threshold depends on the costs of \define{false positives} (an anomaly is reported when the observation is normal) and \define{false negatives} (no anomaly is reported when the observation is anomalous).

\begin{Definition}[Classifier/Predictor]
	A measurable function $\eta\colon (\mathcal{X},\mathscr{X})\to(\mathcal{Q},\mathscr{Q})$ is called a \define{classifier} or \define{predictor}.
\end{Definition}

A classifier calculates a prediction for an observation $X=x$. In the context of anomaly detection our goal is to find a good predictor which can distinguish normal from anomalous data. 
However, the distribution $\PP_X \otimes \PP_Y$ is typically unknown and hence we have to build a classifier solely based on observations. The estimation of such a functional relationship between the input space $\mathcal{X}$ and the prediction space $\mathcal{Q}$ is called \define{learning} or \define{training}. 

\begin{Definition}[Batch Anomaly Detection]
	In \emph{unsupervised batch learning} we have access to $n\in\Nat$ unlabeled independent realizations $(x_1,\dotsc,x_n)$, identically distributed according to $\otimes_{i=1}^n \PP_X$ which form a \emph{training set}.
	In anomaly detection we make the following assumptions:
	\begin{itemize}
		\item The objective is to estimate a predictor $\eta$ based on an unlabeled training set.
		\item The training set contains mostly normal instances from $\PP_{\condition{X\given Y=\CN}}$ and only a few anomalies as, by definition, anomalies are rare events.
		\item It is assumed that the algorithm has complete access to all $n$ elements of the dataset at once.
		\item We may have access to a small labeled fraction of the training data to configure our algorithm.
	\end{itemize}	
\end{Definition}

\begin{Definition}[Online \& Streaming Anomaly Detection]
	In contrast to batch learning, where the full dataset $(x_1,\dotsc,x_n)$ is permanently available, \emph{online} learning algorithms observe each $x_t$, $(1\leq t \leq n)$ only once and in a sequential order. Typically these algorithms have limited memory and thus can only store a small set of previously observed samples. Hence it is necessary to continuously update the prediction function based on the new observations
	\begin{align*}
	(x_t,\eta_t) \mapsto \eta_{t+1}
	\end{align*}
	to build a new predictor. This can be generalized to \emph{streaming anomaly detection} where data arrives in a possible infinite sequence $x_1,x_2,x_3,\dotsc$ of observations. Moreover the input and label space distributions may evolve over time, a problem known as \emph{concept drift}. (Formally we now have to consider the stochastic processes $\SetDef{X_t}_{t\in\Nat}$, $\SetDef{Y_t}_{t\in\Nat}$ and their corresponding distributions).
	It is therefore necessary that an algorithm can adapt to the changes \eg by forgetting outdated information while incorporating new knowledge \autocite{gama2014survey}. 
	These classes of algorithms assume that more recent observations carry more relevant information than older data. 
	In summary streaming anomaly detection has the following key characteristics:
\begin{itemize}
	\item The data stream is possible infinite which requires the algorithm to learn incrementally since it is not possible to store the whole stream.
	\item Most instances in the data stream belong to the class of normal data and anomalies are rare.
	\item The stream can evolve over time, forcing algorithms to adapt to changes in the data distribution.
	\item Only a small time frame at the beginning of the stream is available to configure the algorithm's parameter.
	\item We have to instantly make a prediction $\eta_t(x_t)$ as soon as an observation $x_t$ is available. This requires that predictions can be made fast. 
\end{itemize}	
\end{Definition}

\section{Related Work}\label{sec:related_work}
Many approaches from statistics and machine learning can be used for anomaly detection \citep{Chandola2009,Gupta2014}, but only a few are applicable on high-dimensional, large-scale problems, where a vast amount of information has to be processed. We review several algorithms with focus on their computational complexity and memory requirements.

\subsection{Distribution Based Models}
One of the oldest, statistical methods for anomaly detection is the kernel (or Parzen) density estimator (\acronym{KDE}). With $\BigO(n)$ time for predictions the \acronym{KDE} is too slow for large amounts of data and known to be problematic in the case of increasing data dimensionality \citep{Gretton2012}. 
Fitting parametric distributions such as the Normal, Gamma, etc. is problematic since in general, the underlying data distribution is unknown. Therefore, a mixture of Gaussians is often used as a surrogate for the true distribution as, for example, done by SmartSifter \citep{yamanishi2004line}. SmartSifter can handle multivariate data with both, continuous and categorical observations. The main disadvantage of this approach is the high number of parameters required for the mixture model which grows quadratically with the dimension \citep{Tax2001}.

\subsection{Distance Based Models}

Distance based models are popular since most of them are easy to implement and interpret.
Knorr et al. \citep{knorr1998algorithms,knorr2000distance} labels an observation as a \emph{distance based outlier} (anomaly) if at least a fraction of points in the dataset have a distance of more than a threshold (based on the fraction) to this point.
The authors proposed two simple algorithms which have both $\BigO(n^2)$ runtime and a cell-based version which runs linear in $n$, but exponential with the dimension $d$.  \Citet{ramaswamy2000efficient} argues that the threshold can be difficult to determine and proposes an \emph{outlier score} which is simply the distance from a query point to its $k$th nearest neighbor. The algorithm is called \textsc{KNN}Outlier and suffers from the problem of efficient nearest neighbor search. If the input space is of low dimension and $n$ is much larger than $2^d$ then finding $1$ nearest neighbor in a $k$-$d$ tree with randomly distributed points takes $\BigO(\log n)$ time on average. However this does not hold in high dimensions, where such a tree is not better than an exhaustive search with $\BigO(n)$ \citep{Goodman2004}. Also the algorithm proposed by Ramaswamy et al. is only used to identify the top outliers in a given dataset. An alternative algorithm was proposed by Angiulli et al. \citep{angiulli2002fast} using the sum of distances from its $k$-nearest neighbors. \citet{ott2014integrated} simultaneously perform clustering and anomaly detection in an integer programming optimization task. 

Popular approaches from data mining for distance based novelty detection on streams are \acronym{OLINDDA} \citep{spinosa2007olindda} and its extension \acronym{MINAS} \citep{faria2013novelty} which both represent normal data as a union of spheres obtained by clustering. This representation becomes problematic if data within one cluster exhibits high variance since then the decision boundary becomes too large to detect novelties. Both algorithms are designed to incorporate novel classes into their model of normal data and hence barely applicable to anomaly detection. %

The STream OutlieR Miner (\acronym{STORM}) \citep{angiulli2010distance,angiulli2007detecting} offers an efficient solution to the problem of distance-based outlier detection over windowed data streams using a new data structure called Indexed Stream Buffer. 
Continuous Outlier Detection (\acronym{COD}) \citep{kontaki2011continuous} aims to further improve the efficiency of \acronym{STORM} by reducing the number of range queries.

\subsection{Density Based Models}

Nearest neighbor data description \citep{Tax2001} approximates a local density while using only distances to its first neighbor. The algorithm is very simple and often used as a baseline. It is also relatively slow approaching $\BigO(n)$ per prediction.
More sophisticated is the local density based approach called Local Outlier Factor (\acronym{LOF}) \citep{Breunig2000}. It considers a point to be an anomaly if there are only \emph{relatively} few other points in its neighborhood. \acronym{LOF} was extended to work on data streams \citep{Pokrajac2007}, however both (the batch and incremental approach) are relatively slow with training time between $\BigO(n \log n)$ and $\BigO(n^2)$ and $\BigO(n)$ memory consumption.

The angle based outlier detection for high-dimensional data (\acronym{ABOD}) proposed by Kriegel and Zimek \citep{kriegel2008angle} is able to outperform \acronym{LOF}, however requires $\mathcal{O}(n^2)$ time per prediction with the exact model and $\BigO(n+k^2)$ if the full dataset is replaced by the $k$-nearest neighbors of the query point (FastAbod).

\subsection{Classification \& Tree Based Models}

The One-class support vector machine (\acronym{OC-SVM}) \citep{Scholkopf2001,Tax2004} is a kernel based method which attempts to find a hyperplane such that most of the observations are separated from the origin with maximum margin. This approach does not scale very well to large datasets where predictions have to be made with high frequency. As, \citet{Steinwart2003} showed that the number of support vectors can grow linearly in the size of the dataset. There exist One-class support vector machines which can be learned incrementally \citep{gretton2003line}.

Hoeffding Trees \citep{domingos2000mining} are anytime decision trees to mine high-speed data streams. The Hoeffding Trees algorithm is not applicable to solve the unsupervised anomaly detection problem considered in this work since it requires the availability of class labels.
Streaming Half-Space-Trees (\acronym{HSTa}) \citep{tan2011fast} randomly construct a binary tree structure without any data. It selects a dimension at random and splits it in half. Each tree then counts the number of instances from the training set at each node referred to as \enquote{mass}. The score for a new instance is then proportional to the mass in the leaf in which new instance hits after passing down the tree. Obviously, an ensemble of such trees can be built-in constant time and the training is linear in $n$. However, randomly splitting a very high-dimensional space will not yield in a tree sufficiently fine-grained for anomaly detection. The RS-Forest \citep{wu2014rs} is a modification of \acronym{HSTa} in which each dimension is not splitted in half, but at a random cut-point. Also the assumption that \textcquote{wu2014rs}{\textelp{} once each instance is scored, streaming RS-Forest will receive the true label of the instance \textelp{}} does not always hold.
The Isolation Forest (\textit{i}Forest) is an algorithm which uses a tree structure to isolate instances \citep{liu2012isolation}. The anomaly score is based on the path length to an instance. \textit{i}Forests achieve a constant training time and space complexity by sub-sampling the training set to a fixed size.
The characteristics of the most relevant anomaly detection algorithms is summarized in \cref{tbl:algorithm_comparison}. All complexities are given with respect to the dataset size $n$ in high-dimensional spaces.

Most methods discussed do not scale to very large problems since either the training time is non-linear with the number of samples or the time to make a single prediction increases with the dataset size (stream length). We now present a novel anomaly detection algorithm to overcome these problems.

\begin{table}[tb]
	\caption{Comparison of anomaly detection techniques}
	\small
\begin{tabu}{X[1.2,l]X[c]X[c]X[c]X[.6,c]X[.6,c]X[1,c]X[1.5,l]} 
	\toprule 
	 & training & prediction & memory  & batch & online & streaming & problem size \\ 
	%\midrule
	\cmidrule(lr){2-4}\cmidrule(lr){5-7}\cmidrule(lr){8-8}
	\addlinespace
	\EXPoSE & $\BigO(n)$ & $\BigO(1)$ & $\BigO(1)$ & \cmark & \cmark & \cmark & large \\
	\acronym{OC-SVM} & $\BigO(n^2)$ & $\BigO(n)$ & $\BigO(n)$ & \cmark & \cmark & \xmark & medium\\
	\acronym{LOF} & $\BigO(n^2)$ & $\BigO(n)$ & $\BigO(n)$ & \cmark & \cmark & \xmark & medium\\
	\acronym{KDE} & $\BigO(1)$ & $\BigO(n)$ & $\BigO(n)$ & \cmark & \cmark & \xmark & small\\
	FastAbod & $\BigO(n)$ & $\BigO(n)$ & $\BigO(n)$ & \cmark & \xmark & \xmark & small\\
	\textsc{\textit{i}Forest} & $\BigO(1)$ & $\BigO(1)$ & $\BigO(1)$ & \cmark & \xmark & \xmark & large\\	
	\acronym{STORM} & $\BigO(n)$ & $\BigO(1)$ & $\BigO(1)$ & \xmark & \xmark & \cmark & small \\
	\acronym{COD} & $\BigO(n)$ & $\BigO(1)$ & $\BigO(1)$ & \xmark & \xmark & \cmark & small  \\
	\acronym{HSTa} & $\BigO(n)$ & $\BigO(1)$ & $\BigO(1)$ & \xmark & \xmark & \cmark & medium \\
	\addlinespace
	\bottomrule 
\end{tabu}
\label{tbl:algorithm_comparison}
\end{table}

\section{Expected Similarity Estimation}\label{sec:expose}

As before, let $X$ be a random variable taking values in a measurable space $(\mathcal{X},\mathscr{X})$. We are primarily interested in the distribution of normal data $\PP_{\condition{X\given Y=\CN}}$ for which we will simply use the shorthand notation $\PP$ in the remainder of this work. Next we introduce some definitions which are necessary in the following.

A Hilbert space $(\mathcal{H},\ip{\cdot,\cdot})$ of functions $f\colon\mathcal{X}\to\Real$ is said to be a \define{reproducing kernel Hilbert space} (\RKHS) if the evaluation functional $\bar{\delta}_x\colon f \mapsto f(x)$ is continuous. A function $k\colon \mathcal{X}\times\mathcal{X} \to \Real$ which satisfies the reproducing property
\begin{align*}
\ip{f,k(x, \cdot)} &= f(x) \quad \text{and in particular} \\
\ip{k(x, \cdot),k(y, \cdot)} &= k(x,y)
\end{align*} 
is called \define{reproducing kernel} of $\mathcal{H}$~\autocite{steinwart2008support}.\footnote{The notation $k(x, \cdot)$ indicates that the second function argument is not bound to a variable.}
The map $\phi: \mathcal{X} \to \mathcal{H}$,  $\phi\colon x \mapsto k(x,\cdot)$ with the property that
\begin{align*}
k(x,y) = \ip{\phi(x),\phi(y)}
\end{align*} 
is called \define{feature map}.

Throughout this work we assume that the reproducing kernel Hilbert space $(\mathcal{H},\ip{\cdot,\cdot})$ is \emph{separable} such that $\phi$ is measurable. We therefore assume that the input space $\mathcal{X}$ is a separable topological space and the kernel $k$ on $\mathcal{X}$ is continuous, which is sufficient for $\mathcal{H}$ to be separable \autocite[Lemma 4.33]{steinwart2008support}.

As mentioned in the introduction, \EXPoSE calculates a score which can be interpreted as the likelihood of an instance $z\in\mathcal{X}$ belonging to the distribution of normal data $\PP$. It uses a kernel function $k$ to measure the similarity between instances of the input space $\mathcal{X}$.
\begin{Definition}[Expected Similarity Estimation]
	The \emph{expected similarity} of $z \in \mathcal{X}$ with respect to the (probability) distribution $\PP$ is defined as
	\begin{align*}
	\eta(z) = \ExpOp\left[\phi(z)\right] = \int_{\mathcal{X}} k(z,x) \,\mathrm{d}\PP(x), 
	\end{align*}
	where $k\colon\mathcal{X} \times \mathcal{X} \to \Real$ is a reproducing kernel.
\end{Definition}
Intuitively the query point $z$ is compared to all other points of the distribution $\PP$. We will show that this equation can be rewritten as an inner product between the feature map $\phi(z)$ and the \emph{kernel mean map} $\mu[\PP]$ of $\PP$. This reformulation is of central importance and will enable us to efficiently compute all quantities of interest.
Given a reproducing kernel $k$, the kernel mean map can be used to embed a probability measure into a \RKHS where it can be manipulated efficiently. It is defined as follows.
\begin{Definition}[Kernel Embedding]
	Let $\PP$ be a Borel probability measure on $\mathcal{X}$.
	The \emph{kernel embedding} or \emph{kernel mean map} $\mu[\PP]$ of $\PP$ is defined as
	\begin{align*}
		\mu[\PP] &= \int_{\mathcal{X}} k(x,\cdot) \,\mathrm{d}\PP(x), 
	\end{align*}
	where $k$ is the associated continuous, bounded and positive-definite kernel function.
\end{Definition}
We assume that the kernel $k$ is bounded in expectation \ie 
\begin{align*}
\int_\XX \sqrt{k(x,x)} \dd \PP(x) < \infty, 
\end{align*}
such that $\mu[\PM]$ exists for all Borel probability measures $\PP$ \autocite[Page 8]{sejdinovic2013equivalence}. This is a weaker assumption than $k$ being bounded. We can now continue to formulate the central theorem of our work.
\begin{Theorem}\label{thm:expose_innerprod}
	Let $(\mathcal{H},\ip{\cdot,\cdot})$ be a \RKHS with reproducing kernel $k\colon \mathcal{X}\times\mathcal{X} \to \Real$.
	The expected similarity of $z \in \mathcal{X}$ with respect to the distribution $\PP$ can be expressed as
	\begin{align*}
	\eta(z) &= \int_{\mathcal{X}} k(z,x) \,\mathrm{d}\PP(x) \\
	&= \ip{ \phi(z), \mu[\PP]},
	\end{align*}
	where $\mu[\PP]$ is the kernel embedding of $\PP$.
\end{Theorem}

This reformulation has several desirable properties. At this point we see how the \EXPoSE classifier can make prediction in \emph{constant} time. After the kernel mean map $\mu[\PP]$ of $\PP$ is learned, \EXPoSE only needs to calculate a single inner product in $\HH$ to make a prediction. However there are some crucial aspects to consider \ie 
in Hilbert spaces, integrals and continuous linear forms are not in general interchangeable. 
In the proof of \cref{thm:expose_innerprod} we will thus use the weak integral and show that it coincides with the strong integral. 
\begin{Definition}[Strong Integral]\label{def:bochnerIntegral}
Let $(\XX,\XXX,\PP)$ be a $\sigma$-finite measure space and let $\phi\colon\XX\to\HH$ be measurable. Then $\phi$ is \define{strong integrable} (\define{Bochner integrable}) over a set $\mathcal{D} \in \XXX$ if and only if its norm $\norm{\phi}$ is Lebesgue integrable over $\mathcal{D}$, that is,
\begin{align*}
\int_\mathcal{D}  \norm{\phi} \dd\PP(x) < \infty.
\end{align*}
If $\phi$ is strong integrable over each $\mathcal{D} \in \XXX$ we say that $\phi$ is \emph{strong integrable}. \autocite[Theorem 11.44]{aliprantis2006infinite}
\end{Definition}
\begin{Definition}[Weak Integral]\label{def:weakIntegral}
Let $(\XX,\XXX,\PP)$ be a $\sigma$-finite measure space.
A function $\phi\colon \XX \to \HH$ is \define{weakly integrable} over a set $\mathcal{D} \in \XXX$ if there exists some $\lambda \in \HH$ satisfying
\begin{align*}
\ip{f,\lambda} = \int_\mathcal{D} \ip{f,\phi(x)} \dd\PP(x) 
\end{align*}
for each $f\in\HH$. The weak integral is denoted by
\begin{align*}
\lambda = \oint_\mathcal{D} \phi \dd\PP(x)
\end{align*}
and the unique element $\lambda \in \HH$ is called \define{weak integral} of $\phi$ over $\mathcal{D}$. If the integral exists for each $\mathcal{D} \in \XXX$ we say that $\phi$ is \emph{weakly integrable}. \autocite[Section 11.10]{aliprantis2006infinite}
\end{Definition}
A sufficient condition therefore is provided by the following lemma.

\begin{Lemma}\label{thm:WeakStrongIntegrable}
If $\phi$ is strong (Bochner) integrable then $\phi$ is weak (Pettis) integrable and the two integrals coincide. \autocite[Theorem 11.50]{aliprantis2006infinite}
\end{Lemma}
We are now in the position to proof \cref{thm:expose_innerprod}.
\begin{Proof}[Proof of \cref{thm:expose_innerprod}]
By definition of the feature map $\phi$ we have
\begin{align*}
\int_\XX k(x,z) \dd\PP(x) = \int_\XX \ip{\phi(z),\phi(x)} \dd\PP(x)
\end{align*}
By the assumption that $k$ is bounded in expectation it follows that
\begin{align*}
\int_\XX  \norm{\phi(x)} \dd\PP(x) = \int_\XX  \sqrt{k(x,x)} \dd\PP(x) < \infty
\end{align*}
and therefore $\phi$ is strongly integrable and hence weakly integrable (\cref{thm:WeakStrongIntegrable}).
By definition of the weak integral we get for all $z\in\XX$ 
\begin{align*}
\int_\XX \ip{\phi(z),\phi(x)} \dd\PP(x) &= \ip[\Big]{\phi(z),\oint_\XX \phi(x) \dd\PP(x)} \\
&= \ip[\Big]{\phi(z),\int_\XX \phi(x) \dd\PP(x)}\\
& = \ip{ \phi(z), \mu[\PP]}
\end{align*}
for all probability measures $\PP$. \qed
\end{Proof}

In anomaly detection, we cannot assume to know the distribution of normal data $\PM$. However we assume to have access to $n\in\Nat$ independent realizations $(x_1,\dotsc,x_n)$ sampled from $\PP$. It is common in statistics to estimate $\PP$ with the empirical distribution
\begin{align*}
\PP_n = \frac{1}{n}\sum_{i=1}^n \delta_{x_i},
\end{align*}
where $\delta_{x}$ is the Dirac measure.
The empirical distribution $\PP_n$ can also be used to construct an approximation $\mu[\PP_n]$ of $\mu[\PP]$ as
\begin{align*}
\mu[\PP] \approx  \mu[\PP_n]  = \frac{1}{n}\sum_{i=1}^n \phi(x_i)
\end{align*}
which is called \emph{empirical kernel embedding}~\autocite{Smola2007}. 
This is an efficient estimate since it can be shown \autocite{Schneider2016} that under the assumption $\norm{\phi(X)} \leq c$ with $c>0$ the difference between $\muP$ and $\mun$ is in probability
\begin{align*}
\Pr\left( \norm[\big]{\muP - \mun} \geq \epsilon \right) 
\leq 2 \exp\bigg( -\frac{n\epsilon^2}{8c^2} \bigg) 
\end{align*}
for all $\epsilon > 0$. 

As a consequence we can substitute $\muP$ with $\mun$ whenever the distribution $\PP$ is not directly accessible yielding
\begin{align*}
\eta(z) &= \ip{\phi(z),\mun} \\
& = \ip[\Big]{\phi(z),\frac{1}{n}\sum_{i=1}^{n}\phi(x_i)}
\end{align*}
as the (empirical) \EXPoSE anomaly detector. The empirical kernel embedding $\mu[\PP_n]$ is responsible for the \emph{linear} training computational complexity of \EXPoSE. We will call $\mu[\PP_n]$ the \EXPoSE \emph{model}.
One of the important observations is, that \EXPoSE makes no assumption about the type or shape of the data distribution $\PP$ as such assumption can be wrong, causing erroneous predictions. This is an advantage over other statistical approaches that try to approximate the distribution directly with parametric models. 

\subsection{Parallel \& Distributed Processing}

Parallel and distributed data processing is the key to scalable machine learning algorithms. The formulation of \EXPoSE as $\eta(z) = \ip{\phi(z),\mun}$ is especially appealing for this kind of operations. We can use a \acronym{SPMD} (single program, multiple data) technique to achieve parallelism. On of the first programming paradigms on this line is  Google’s \emph{MapReduce} for processing large data sets on a cluster \autocite{dean2008mapreduce}.

Assume a partition of the dataset $(x_1,\dotsc,x_n)$ into $m \leq n$ distinct collections $s_1,\dotsc,s_m$ which can be distributed on different computational nodes.
Obviously, the feature map $\phi$ can be applied in parallel to all instances in $x_1,\dotsc,x_n$.
We also note that the partial sums 
\begin{align*}
p(s_i) = \sum_{x \in s_i} \phi(x)
\end{align*}
can be calculated without any communication or data-sharing across concurrent computations.
Solely the partial sums $p(s_i)$, which are elements of $\HH$, need to be transmitted and combined as
\begin{align*}
\mun = \frac{1}{n}\sum_{i=1}^{m} p(s_i)
\end{align*}
by a central processing node.

\subsubsection*{Summary}

In this section we derived the \EXPoSE anomaly detection algorithm. We showed how \EXPoSE can be expressed as an inner product $\ip{\phi(z),\mun}$ between the kernel mean map of $\PP$ and the feature mapping of a query point $z\in\mathcal{X}$ for which we need to make a prediction. Evaluating this inner product takes \emph{constant} time while estimating the model $\mun$ can be done in \emph{linear} time and with \emph{constant} memory. We will explain the calculation of $\phi$ in more detail in \cref{sec:featuremaps} and will explore now how \EXPoSE can be learned incrementally and applied to large-scale data streams.

\section{Online \& Streaming \EXPoSE}\label{sec:oexpose}

In this section we will show how \EXPoSE can be used for online and streaming anomaly detection. To recap, a \emph{data stream} is an often infinite sequence of observations $(x_1,x_2,x_3,\dotsc)$, where $x_t \in \mathcal{X}$ is the instance arriving at time $t$. A source of such data can be, for example, continuous sensor readings from an engine or a video stream from surveillance cameras. 

\Citet{domingos2001catching} identified the following requirement for algorithms operating on \enquote{the high-volume, open-ended data streams we see today}.
\begin{itemize}
\item Require small constant time per instance.
\item Use only a fixed amount of memory, independent of the number of past instances.
\item Build a model using at most one scan over the data.
\item Make a usable predictor available at any point in time.
\item Ability to deal with concept drift.
\item For streams without concept drift, produce a predictor that is equivalent (or nearly identical) to the one that would be obtained by an offline (batch) learning algorithm.
\end{itemize}

In this section we will show that the online version of \EXPoSE fulfills all requirements, starting with the last item of the list. 

\begin{Proposition}\label{thm:incremental_expose}
	The \EXPoSE model $\mun$ can be learned incrementally, where each model update can be performed in $\mathcal{O}(1)$ time and memory.
\end{Proposition}
\begin{Proof}
	Given a stream $(x_1,x_2,x_3,\dotsc)$ of observations and let $\mu[\PM_1] = \phi(x_1)$.
	Whenever a new observation $x_{t}$ is made at $t > 1$, the new model $\mu[\PM_t]$ can be incrementally calculated as
	\begin{align*}
	\mu[\PM_t] &= \frac{1}{t} \sum_{i=1}^{t} \phi(x_i) \nonumber\\
	&= \mu[\PM_{t-1}] + \frac{1}{t} \Big(\phi(x_t) - \mu[\PM_{t-1}] \Big)
	\end{align*} 
	using the previous model $\mu[\PM_{t-1}]$.
\end{Proof}
We see that online learning of \EXPoSE does neither increase the computational complexity nor the memory requirements of \EXPoSE.
We also emphasize that online learning yields the exact same model as the \EXPoSE offline learning procedure.

\subsection{Learning on Evolving Data Streams}
\begin{figure*}[tb] 
	\begin{center}
	\includegraphics{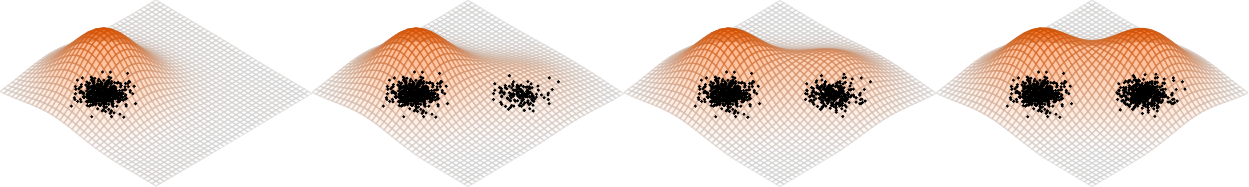}
	\includegraphics{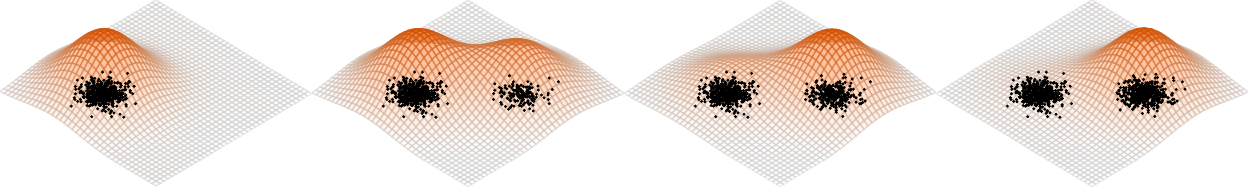}
		\caption{An illustration of the difference between online (incremental) learning and model adaption to concept drift. The eight plots represent \EXPoSE predictions for the observations indicated by the black dots. Each row displays four snapshots with increasing time from left to right as data becomes available from two clusters. We first sampled only from the left cluster and later only from the right.
			In the top row we incrementally build the model by adding more knowledge to it, whereas in the bottom row the model evolves and slowly forgets outdated observations (the points from the left cluster).}
		\label{fig:cmp_incremental_drift}
	\end{center}
\end{figure*}

Sometimes it can be expected that the underlying distribution of the stream evolves over time. This is a property known as \emph{concept drift} \citep{sadik2014research}. For example in environmental monitoring, the definition of \enquote{normal temperature} changes naturally with seasons. We can also expect that human behavior changes over time which requires us to redefine what anomalous actions are.
In \cref{fig:cmp_incremental_drift} we illustrate the difference between incremental learning as in \cref{thm:incremental_expose} and a model which adapts itself to changes in the underlying distribution. In the following we will use $w_t$ to denote the \EXPoSE model at time $t$ since the equation $w_t=\mu[\PM_t]$ will \emph{not} necessarily hold when concept drift adaptation is implemented.

In this work we are not concerned with the detection of concept drift \citep{gama2010knowledge}, but we will show how \EXPoSE can be used efficiently with the most common approaches to concept drift adaption which either utilize \emph{windowing} or \emph{forgetting} mechanisms \citep{gama2014survey}.

\subsubsection{Windowing}\label{sec:windowing}

Windowing is a straight forward technique which uses a buffer (the window) of $l\in\Nat$ previous observations. Whenever a new observation is added to the window, the oldest one is discarded. We can efficiently implement windowing for \EXPoSE as follows.

\begin{Proposition}\label{thm:window_expose}
	Concept drift adaption on data streams using a sliding window mechanism can be implemented for \EXPoSE with $\mathcal{O}(1)$ time and $\mathcal{O}(l)$ memory consumption, where $l\in\Nat$ is the window size.
\end{Proposition}
\begin{Proof}
	Given a data stream $(x_1,x_2,x_3\dotsc)$ and the window size $l$. For $t < l$ we set $w_t = \frac{1}{t} \sum_{i=1}^{t}$ and use
	the incremental update
	\begin{align*}
	w_t &= \frac{1}{l} \sum_{i=t-l+1}^{t} \phi(x_i) \nonumber\\
	&= w_{t-1} + \frac{1}{l}\phi(x_t) - \frac{1}{l}\phi(x_{t-l}), %\label{eq:sliding_windows}
	\end{align*} 
	whenever $t \geq l$.
\end{Proof}

The downside of a sliding window mechanism is the requirement to keep the past $l\in\Nat$ events in memory. Also the sudden discard of a data point can lead to abrupt changes in predictions of the classifier which is sometimes not desirable. 
Another question is how to choose the correct window size. A shorter sliding window allows the algorithm to react faster to changes and requires less memory though the available data might not be representative or noise has too much negative impact. On the other hand a wider window may take too long to adapt to concept drift. The window size is therefore often dynamically adjusted \citep{widmer1996learning} or multiple competing windows are used \citep{lazarescu2004using}.

\subsubsection{Gradual Forgetting (Decay)}\label{sec:gradualfrogetting}

The problems of sliding window approaches can be avoided if a forgetting mechanism is applied, where the influence of older data gradually vanishes. Typically a parameter can be used to control the tradeoff between fast adaptation to new observations and robustness against noise in the data. We can realize such a forgetting mechanism for \EXPoSE by replacing the factor $\frac{1}{t}$ in \cref{thm:incremental_expose} by a constant $\gamma \in [0,1)$ yielding
\begin{align*}
w_{t} =%&= \gamma \phi(x_t) - (1-\gamma) \mu[\PM_{t-1}] 
  \begin{cases}
  \phi(x_t) & \text{for } t= 1 \\
  \gamma \phi(x_t) + (1-\gamma) w_{t-1}, & \text{for } t>1
  \end{cases}
\end{align*} 
where, with $\gamma=0$, no new observations are integrated into the model. This operation can be performed in constant time as summarized in the next proposition.

\begin{Proposition}\label{thm:decay_expose}
	Concept drift adaptation on data streams using a forgetting mechanism can be implemented for \EXPoSE in $\mathcal{O}(1)$ time and memory.
\end{Proposition}
\begin{Proof}
	This is a direct consequence from \cref{thm:incremental_expose}.
\end{Proof} 

In general, weighting with a fixed $\gamma$ or using a static window size is called \emph{blind} adaptation since the model does not utilize information about changes in the environment \citep{gama2010knowledge}. The alternative is \emph{informed} adaptation where one could, for example, use an external change detector \citep{gama2014survey} and weight new samples more if a concept drift was detected. We could also apply more sophisticated decay rules making $\gamma$ a function of $t$ or $x_t$.

A summary of characteristics for each proposed online learning variant of \EXPoSE is listed in \cref{tbl:onlinelearning} and a general discussion can be found in literature, \eg the work of \citet{gama2010knowledge}.

\begin{table}
\begin{center}
	\caption{Comparison of online learning techniques for \EXPoSE}
	\small
	\begin{tabular}{lp{45mm}p{45mm}} 
		\toprule 
		\addlinespace
		& \multicolumn{1}{c}{\textsc{Pros}}  & \multicolumn{1}{c}{\textsc{Cons}} \\ 
		\cmidrule(lr){2-2}\cmidrule(lr){3-3}
		\addlinespace
		Prop. \ref{thm:incremental_expose}: online  &% 
			\cmark~equivalent to batch version
		&%
			\xmark~no concept drift adaptation 
		\\
		Prop. \ref{thm:window_expose}: window  &%
			\cmark~concept drift adaptation
		 &%
			\xmark~possibly sudden changes in predictions\newline 
			\xmark~difficult to choose window size\newline 
			\xmark~increased memory requirements for window buffer\\
		Prop \ref{thm:decay_expose}: decay  &%
		\cmark~concept drift adaptation\newline
		\cmark~gradual vanishing influence of outdated information\newline
		\cmark~no increased memory requirements
		&\\
		\addlinespace
		\bottomrule
	\end{tabular}
	\label{tbl:onlinelearning}
\end{center}
\end{table}

\subsection{Predictions on Data Streams}

We introduced three different approaches to learn the model for \EXPoSE on data streams. One incremental (online) learning approach and two evolving techniques. In order to make a prediction as a new observation is made we have to normalize the calculated predicted score. This is necessary as the score would continuously change, even if exactly the same data would be observed again. This problem is not present in the batch version of \EXPoSE since the model does not change anymore at the time we make predictions. To avoid this problem we divide by the total volume
\begin{align*}
\int_{\mathcal{X}}\int_{\mathcal{X}} k(x,y) \,\mathrm{d}\PP(x)\mathrm{d}\PP(y) &= \innerprod{\muP}{\muP}\\
&\approx \innerprod{w_t}{w_t}
\end{align*}
yielding
\begin{align*}
\eta(z) = \frac{\innerprod{\phi(z)}{w_t}}{\norm{w_t}^2},
\end{align*}
as the \EXPoSE classifier.
We emphasize that the calculation of the normalization constant does not change the limiting behavior of runtime and memory we derived earlier in this section since we have constant time access to $w_t$ anyway.

\section{Approximate Feature Maps}\label{sec:featuremaps}

We showed in the previous part how \EXPoSE can be expressed as an inner product $\ip{\phi(z),\mun}$ between the kernel mean map and the feature map of a query point $z\in\mathcal{X}$ and derived a similar expression for the incremental and streaming variants of \EXPoSE. 
However, the feature map $\phi$ (and hence $\mun$) can not always be calculated explicitly as $\phi(z) = k(\cdot,z)$. One possible solution is to resort to approximate feature maps which we review in this section. The key idea behind approximate feature maps is to find a function $\hat{\phi}$ such that
\begin{align*}
k(x,z) \approx \innerprod{\hat{\phi}(x)}{\hat{\phi}(z)}
\end{align*}
and $\hat{\phi}(x) \in \Real^r$ for some $r\in\Nat$. We will see, that this can be done efficiently.

\subsection{Random Kitchen Sinks}

%\tikzexternaldisable
\begin{figure}[tb]
	\includegraphics{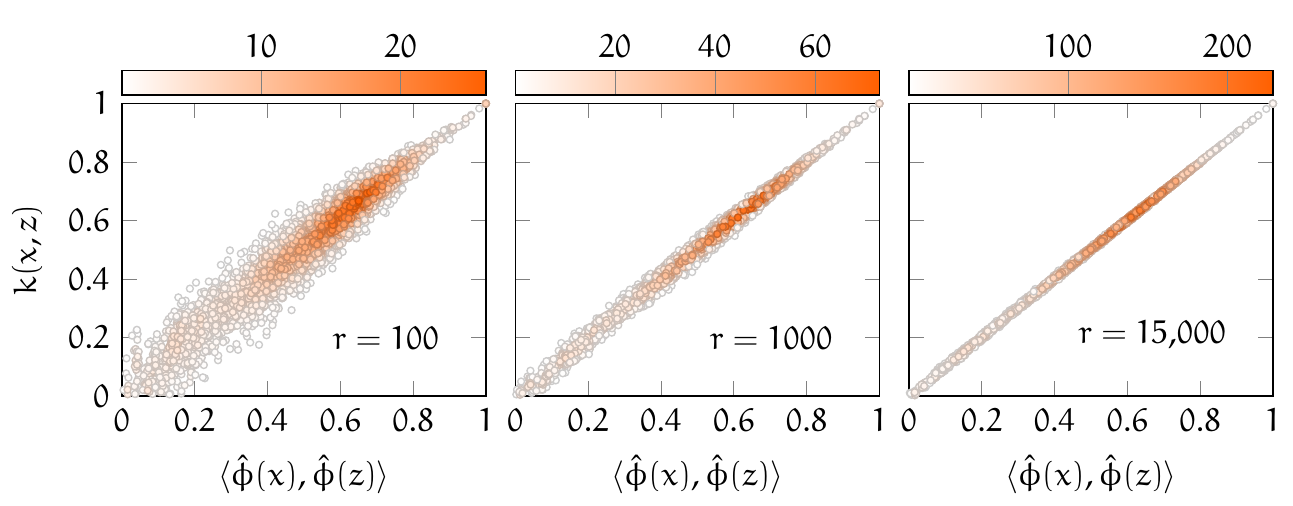}
	\caption{A comparison between the values calculated by a Gaussian \acronym{RBF} kernel $k(x,z)$ and the \acronym{RKS} approximation $\innerprod{\hat{\phi}(x)}{\hat{\phi}(z)}$ for random images of the \acronym{MNIST} dataset. The individual plots show how the number of kernel expansions $r$ affect the approximation quality. Color indicates the density.}
	\label{fig:rks_approx_kernel}
\end{figure}

A way to efficiently create a feature map $\phi$ from a kernel $k$ is known as \emph{random kitchen sinks} \citep{Rahimi2007,Rahimi2008}.
The random kitchen sinks (\acronym{RKS}) approximation is based on Bochner's theorem for translation invariant kernels (such as Laplace, Mat\'ern, Gaussian \acronym{RBF}, etc.) and states that such a kernel can be represented as 
\begin{align*}
k(x,y) &= \int_z \phi^\star_z(x) \phi_z(y) \lambda(z) \text{ with } \phi_z(x) = \euler^{\im \innerprod{z}{x}},% \label{eq:bochner}
\end{align*}
where $\phi^\star$ is the conjugate transpose of $\phi$. A Monte Carlo approximation of this integral can then be used to estimate the expression above as 
\begin{align*}
k(x,y) \approx \frac{1}{r}\sum_{i=1}^r \phi^\star_{z_i}(x) \phi_{z_i}(y) = \innerprod{\hat{\phi}(x)}{\hat{\phi}(y)} \quad \text{with} \quad z_i \sim \lambda. 
\end{align*}
For kernels such as the Gaussian \acronym{RBF} $k(x,y) = \exp(-\frac{1}{2}\norm{x-y}^2/\sigma^2)$, the measure $\lambda$ can be found with the help of the inverse Fourier transform yielding
\begin{align*}
\hat{\phi}(x) &= \frac{1}{\sqrt{r}}\exp(\im Zx) \quad \text{with} \quad Z_{ij} \sim \mathcal{N}(0,\sigma^{2}) \text{ and } Z \in \mathbb{R}^{r \times d}
\end{align*}
where $d$ is the input space dimension. The parameter $r \in \Nat$ determines the number of kernel expansions and is typically around \num{20000}. 
Larger $r$ result in better kernel approximations as the Monte Carlo approach becomes more accurate (\cref{fig:rks_approx_kernel}).
Recently \citet{Le2013} proposed an approximation of $Z$ such that the product $Zx$ can be calculated in $\mathcal{O}(r\log d)$ time complexity while requiring only $\mathcal{O}(r)$ storage.

\subsection{Nystr\"om's Approximation}

An alternative to random kitchen sinks are Nystr\"om methods \citep{williams2001using} which project the data into a subspace $\mathcal{H}_r \subset \mathcal{H}$ spanned by $r \leq n$ randomly chosen elements $\phi(x_1), \dotsc, \phi(x_r)$. 

The Nystr\"om feature map $\hat{\phi}$ is then given by $\hat{\phi}(x) = (\hat{\phi}_1(x),\dotsc,\hat{\phi}_r(x))$ with
\begin{align*}
\hat{\phi}_i(x) = \frac{1}{\sqrt{\lambda_{i}}} \sum_{j=1}^r u_{ji}k(x_j,x),\qquad 1\leq i \leq r,
\end{align*}
where $\lambda_{i}$ and  $u_{i}$ denote the $i$-th eigenvalue and the $i$-th eigenvector of kernel matrix $K\in \Real^{r\times r}$ with $K_{i,j} = k(x_i,x_j)$.

The Nystr\"om approximation needs in general less basis functions, $r$, than the \acronym{RKS} approach (typically around \num{1000}). However the approximation is data dependent and hence becomes erroneous if the underlying distribution changes or when we are not able to get independent samples from the dataset.
This is a problem for online learning and streaming applications with concept drift. We therefore suggest to avoid the Nystr\"om feature map in this context.

Random kitchen sinks and the Nystr\"om approximation the most common feature map approximations. We refer to the corresponding literature for a discussion of other approximate feature maps such as \autocite{li2010random,vedaldi2012efficient,kar2012random}, which can be used as well for \EXPoSE.

\subsection{\EXPoSE \& Approximate Feature Maps}
Recall from the previous sections that \EXPoSE uses the inner product $\ip{\phi(z),\mun}$ to calculate the score and make predictions. Using an approximate feature map $\hat{\phi}$, it is now possible to explicitly represent the feature function and consequently also the mean map $\mun$ as
\begin{align}
\eta(z) \approx \innerprod[\Big]{\hat{\phi}(z)}{\frac{1}{n} \sum_{i=1}^n \hat{\phi}(x_i)} \label{eq:approx_mean_map_feat} 
\end{align}
for the \EXPoSE classifier.

We emphasize that with an efficient approximation of $\phi$, as showed here, the training time of this algorithm is \emph{linear} in the number of samples $n$ and an evaluation of $\eta(z)$ for predictions takes only \emph{constant time}. Moreover we need only $\mathcal{O}(r)$ memory to store the model which is also independent of $n$ and the input dimension $d$.

\section{Experimental Evaluation}\label{experiments}

In this section we show in several experiments how \EXPoSE compares to other state of the art anomaly detection techniques in prediction and runtime performances. We first explain which statistical test are used to compare the investigated algorithms.

\subsection{Statistical Comparison of Algorithms}

When comparing multiple (anomaly detection) algorithms over multiple datasets one cannot simply compare the raw numbers obtained from the area under 
\define{receiver operating characteristic} (\acronym{AUC}) or \emph{precision-recall} curves. \Citet{webb2000multiboosting} warns against averaging these numbers: \textcquote{webb2000multiboosting}{It is debatable whether error rates in different domains are commensurable, and hence whether averaging error rates across domains is very meaningful}.

As \citet{demvsar2006statistical} points out, it is also dangerous to use tests which are designed to compare a pair of algorithms for more than two: 
\textcquote{demvsar2006statistical}{A common example of such questionable procedure would be comparing seven algorithms by conducting all 21 paired $t$-tests \textelp{}. When so many tests are made, a certain proportion of the null hypotheses is rejected due to random chance, so listing them makes little sense.}

\citeauthor{demvsar2006statistical} suggests to use the Friedman test
with the corresponding post-hoc Nemenyi test for comparison of more classifiers over multiple data sets. A methodology we summarize in the following.

\subsubsection{The Friedman Test}

The Friedman test \autocite{friedman1937use} is a non-parametric statistical test which ranks algorithms for each dataset individually starting from 1 as the best rank. Its purpose is to examine whether there is a significant difference between the performances of the individual algorithms. Lets assume we compare $k$ algorithms on $m$ datasets and let $r_{ij}$ be the rank of the $j$-th algorithm on the $i$-th dataset. We use $\bar{r}_j$ to denote the average rank of algorithm $j$ given by $\bar{r}_j = m^{-1}\sum_i r_{ij}$. The Friedman statistic 
\begin{align*}
\chi^2_F = \frac{12m}{k(k+1)} \bigg( \sum_{j=1}^{k} \bar{r}_j^2 - \frac{k(k+1)^2}{4} \bigg)
\end{align*}
is undesirably conservative and therefore \citet{iman1980approximations} suggest to use
\begin{align*}
F_F = \frac{(m-1)\chi^2_F}{m(k-1)-\chi^2_F}
\end{align*}
which is distributed according to the $F$-distribution with $k-1$ and ${(k-1){}(m-1)}$ degrees of freedom. If the null-hypothesis (all algorithms are equivalent) is rejected one can proceed with a post-hoc test.

\subsubsection{The Nemenyi Test}

The Nemenyi test \autocite{nemenyi1962distribution} is a post-hoc test to compare all (anomaly detection) algorithms with each other. Hereby the performance of two algorithms is significantly different if their average ranks differ by at least
\begin{align*}
\acronym{CD} = q_\alpha \sqrt{\frac{k(k+1)}{6m}},
\end{align*}
called the \define{critical difference}. Here $q_\alpha$ is the Studentised range statistic divided by $\sqrt{2}$.

%\tikzexternaldisable
\begin{figure}[tb]
	\centering
	\includegraphics{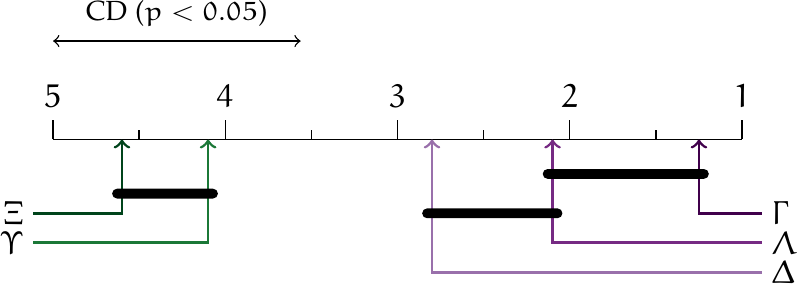}
	\caption[Nemenyi test]{Visualization of the post-hoc Nemenyi test in form of a critical difference diagram. The position on the line indicates the algorithms's average rank. Algorithms which are not significantly different (at $p < 0.05$) are connected with a bar. The critical difference (\acronym{CD}) is given by the double arrow in the top left.}
	\label{fig:critical_difference_example}
\end{figure}
%\tikzexternalenable

\Citet{demvsar2006statistical} also suggests to visually represent the results of the Nemenyi test in a critical difference diagram as in \cref{fig:critical_difference_example}. In this diagram we compare 5 algorithms on 20 datasets against each other. Algorithms not connected by a bar have a significantly different performance.
	
\subsection{Batch Anomaly Detection}

The aim of this experiment is to compare \EXPoSE against \emph{i}Forest, \acronym{OC-SVM}, \acronym{LOF}, \acronym{KDE} and FastAbod in terms of anomaly detection performance and processing time in a learning task without concept drift. In order to be comparable, we follow \citet{liu2012isolation} and perform an outlier selection task with the objective to identify anomalies in a given dataset.

\subsubsection{Datasets}

For performance analysis and evaluation we take the following datasets which are often used in literature for comparison of anomaly detection algorithms as for example in \citep{Scholkopf2001, Tax2004,liu2012isolation}. We use several smaller benchmark datasets with known anomaly classes such as \emph{Ionosphere}, \emph{Arrhythmia}, \emph{Pima}, \emph{Satellite}, \emph{Shuttle}, \citep{Lichman:2013}, \emph{Biomed} and Wisconsin Breast Cancer (\emph{Breastw}) \citep{Tax2004}. These datasets are set up as described in \citep{liu2012isolation} where all nominal and binary attributes are removed.

The larger datasets are the \textsc{Kdd Cup 99} network intrusion data (\textsc{\textsc{KddCup}}) and Forest Cover Type ({ForestCover}). For \textsc{\textsc{KddCup}} instances we follow the setup of \citep{yu2003classifying} and obtain a total of 127 attributes.
Furthermore, we add two high-dimensional image datasets \acronym{\acronym{MNIST}} and the Google Street View House Numbers (\acronym{\acronym{SVHN}}) \citep{Netzer2011}. We use the scaled version of \acronym{\acronym{MNIST}} \citep{Chang2011} and create \acronym{HOG} features \citep{vondrick2013hoggles} for \acronym{\acronym{SVHN}}.
The methodology suggested by \citep{Scholkopf2001,Tax2001} is used to create anomaly detection datasets from \acronym{MNIST} and \acronym{SVHN} in the following way. We take all images of digit $1$ from \acronym{MNIST} as normal instances.
The images of the remaining digits ($2,3,\dotsc,9$) are used as anomalies. We then create a dataset comprising all normal instances and a random subset anomalies such that anomalies account for $1\%$ of the elements in the set. We repeat this process for \acronym{MNIST} images of digits $2,3,\dotsc,9$ and do the same with the $9$ digit classes of \acronym{SVHN} to create $18$ anomaly detection datasets. The subset of anomalies is independently sampled for each repetition of an experiment.
\Cref{tbl:batch_datasets} provides an overview of the dataset properties and how the anomaly classes are defined.

\begin{table}[p]
	\caption{Batch dataset properties}
\begin{center}
	\small
	\begin{tabular}{lrrlc}
		\toprule 
		\addlinespace
		& \multicolumn{1}{c}{\textsc{size ($n$)}}  & \multicolumn{1}{c}{\textsc{dim. ($d$)}} & \multicolumn{1}{c}{\textsc{$\CA$ (anomaly class)}} & \multicolumn{1}{c}{\textsc{$\CA$ proportion}} \\ 
		\cmidrule(lr){2-2}\cmidrule(lr){3-3}\cmidrule(lr){4-4}\cmidrule(lr){5-5}
		\addlinespace
		\textsc{\textsc{KddCup}} & \num{1036241} & \num{127} & \enquote{attack}& 0.3\%\\
		\textsc{ForestCover} & \num{286048} & \num{10} & class 4 vs. 2 & ~~9\%\\
		\textsc{\acronym{\acronym{MNIST}} 1}	& \num{101968}	& \num{784} & 2,3,4,5,6,7,8,9 & ~~1\%\\
		\textsc{\acronym{\acronym{MNIST}} 2}	& \num{90196}	& \num{784} & 1,3,4,5,6,7,8,9 & ~~1\%\\
		\textsc{\acronym{\acronym{MNIST}} 3}	& \num{92763}	& \num{784} & 1,2,4,5,6,7,8,9 & ~~1\%\\
		\textsc{\acronym{\acronym{MNIST}} 4}	& \num{88417}	& \num{784} & 1,2,3,5,6,7,8,9 & ~~1\%\\
		\textsc{\acronym{\acronym{MNIST}} 5}	& \num{82062}	& \num{784} & 1,2,3,4,6,7,8,9 & ~~1\%\\
		\textsc{\acronym{\acronym{MNIST}} 6}	& \num{89536}	& \num{784} & 1,2,3,4,5,7,8,9 & ~~1\%\\
		\textsc{\acronym{\acronym{MNIST}} 7}	& \num{94771}	& \num{784} & 1,2,3,4,5,6,8,9 & ~~1\%\\
		\textsc{\acronym{\acronym{MNIST}} 8}	& \num{88568}	& \num{784} & 1,2,3,4,5,6,7,9 & ~~1\%\\
		\textsc{\acronym{\acronym{MNIST}} 9}	& \num{90034}	& \num{784} & 1,2,3,4,5,6,7,8 & ~~1\%\\
		\textsc{\acronym{\acronym{SVHN}} 1}	& \num{91475}	& \num{2592} & 2,3,4,5,6,7,8,9 & ~~1\%\\
		\textsc{\acronym{\acronym{SVHN}} 2}	& \num{75466}	& \num{2592} & 1,3,4,5,6,7,8,9 & ~~1\%\\
		\textsc{\acronym{\acronym{SVHN}} 3}	& \num{61376}	& \num{2592} & 1,2,4,5,6,7,8,9 & ~~1\%\\
		\textsc{\acronym{\acronym{SVHN}} 4}	& \num{51135}	& \num{2592} & 1,2,3,5,6,7,8,9 & ~~1\%\\
		\textsc{\acronym{\acronym{SVHN}} 5}	& \num{54034}	& \num{2592} & 1,2,3,4,6,7,8,9 & ~~1\%\\
		\textsc{\acronym{\acronym{SVHN}} 6}	& \num{41965}	& \num{2592} & 1,2,3,4,5,7,8,9 & ~~1\%\\
		\textsc{\acronym{\acronym{SVHN}} 7}	& \num{44438}	& \num{2592} & 1,2,3,4,5,6,8,9 & ~~1\%\\
		\textsc{\acronym{\acronym{SVHN}} 8}	& \num{35709}	& \num{2592} & 1,2,3,4,5,6,7,9 & ~~1\%\\
		\textsc{\acronym{\acronym{SVHN}} 9}	& \num{34793}	& \num{2592	} & 1,2,3,4,5,6,7,8 & ~~1\%\\
		\textsc{Shuttle} & \num{58000} & \num{9} & classes 2,3,4,5,7 & ~~6\%\\		
		\textsc{Satellite} & \num{6435} & \num{36} & classes 2,4,5 & 32\%\\		
		\textsc{Pima} & \num{768} & \num{8} & \enquote{pos} & 35\%\\
		\textsc{Breastw} & \num{683} & \num{9} & \enquote{malignant} & 35\%\\
		\textsc{Arrhythmia} & \num{452} & \num{274} & classes 3,4,5,7,8,9,14,15 & 14\%\\
		\textsc{Ionosphere} & \num{351} & \num{32} & \enquote{bad} & 36\%\\
		\textsc{Biomed} & \num{194} & \num{5} & \enquote{carrier} & 34\%\\
		\addlinespace
		\bottomrule 
	\end{tabular}
\end{center}
\label{tbl:batch_datasets}
\end{table}

In the experiment we provide a dedicated labeled random subset of 1\% or 2000 instances (whichever is smaller) to configure the algorithms parameters. We emphasize that this subset is not used to evaluate the predictive performance. The parameter configuration is done by a pattern search \citep{torczon1997convergence} using cross-validation. 
Examples of parameters being optimized are the number of nearest neighbors in \acronym{LOF} and FastAbod, the kernel bandwidth of \EXPoSE, \acronym{KDE} and \acronym{OC-SVM} or the number of trees for \emph{i}Forest. We do not optimize over different distance metrics and various kernels functions, but use the most common Euclidean distance and squared exponential kernel, respectively. However, we remark that the choice of these functions pose a possibility to include domain and expert knowledge into the system. Each experiment is repeated 5 times and their \acronym{AUC} scores are used to perform the Friedman test. 
If not stated otherwise we use \EXPoSE in combination with Nystr\"om's approximation for batch anomaly detection and random kitchen sinks in the streaming experiments as discussed in \cref{sec:featuremaps}. 

\subsubsection{Evaluation}

The average scores for each experiment are reported in \cref{tbl:AUC:batch}, whereas the runtimes are provided in \cref{tbl:time:batch}. Some algorithms failed on the larger datasets. For example \acronym{LOF} was not able to process the \textsc{KddCup} dataset due to the high memory requirements of the tree data structure. However the advantage of a tree data structure for nearest neighbor lookup in low dimensions can be seen when comparing the runtime of \acronym{LOF} on ForestCover and \acronym{MNIST}. Even though the ForestCover dataset has more than 2.5 times the size of \acronym{MNIST}, it takes only a fraction of the time to be processed. This advantage vanishes in higher dimensions.
\acronym{KDE} and FastAbod exhibit a good anomaly detection performance on small datasets, however fail as soon as we apply them to medium-sized problems.

\begin{table*}
	\centering
	\newcolumntype{H}{>{\hsize=7\hsize}X}
	\newcolumntype{Y}{S[table-format=2,table-number-alignment=center]}
	\caption{Batch anomaly detection performances [\acronym{AUC}]}
	\small
	\begin{tabularx}{.95\linewidth}{HYYYYYY}
	\toprule
		& \multicolumn{1}{c}{\textsc{\EXPoSE}}  & \multicolumn{1}{c}{\textsc{\emph{i}Forest}} & \multicolumn{1}{c}{\acronym{\acronym{OC-SVM}}} & \multicolumn{1}{c}{\acronym{\acronym{LOF}}} & \multicolumn{1}{c}{\acronym{\acronym{KDE}}}  & \multicolumn{1}{c}{\textsc{FastAbod}}  \\ 
		\cmidrule(lr){2-2}\cmidrule(lr){3-3}\cmidrule(lr){4-4}\cmidrule(lr){5-5}\cmidrule(lr){6-6}\cmidrule(lr){7-7}
		\addlinespace
\textsc{\textsc{KddCup}} &\maxf{\num{1.00}} &\num{0.99} &\maxf{\num{1.00}} &$\diamond$ &$\star$ &$\star$\\
\textsc{ForestCover} &\num{0.83} &\num{0.87} &\maxf{\num{0.89}} &\num{0.56} &$\star$ &$\star$\\
\textsc{\acronym{\acronym{MNIST}} 1} &\maxf{\num{1.00}} &\num{0.99} &\maxf{\num{1.00}} &\num{0.97} &$\star$ &$\star$\\
\textsc{\acronym{\acronym{MNIST}} 2} &\num{0.79} &\num{0.70} &\num{0.80} &\maxf{\num{0.85}} &$\star$ &$\star$\\
\textsc{\acronym{\acronym{MNIST}} 3} &\num{0.86} &\num{0.70} &\num{0.80} &\maxf{\num{0.88}} &$\star$ &$\star$\\
\textsc{\acronym{\acronym{MNIST}} 4} &\num{0.88} &\num{0.81} &\maxf{\num{0.93}} &\num{0.87} &$\star$ &$\star$\\
\textsc{\acronym{\acronym{MNIST}} 5} &\maxf{\num{0.89}} &\num{0.69} &\num{0.82} &\maxf{\num{0.89}} &$\star$ &$\star$\\
\textsc{\acronym{\acronym{MNIST}} 6} &\maxf{\num{0.94}} &\num{0.86} &\maxf{\num{0.94}} &\num{0.89} &$\star$ &$\star$\\
\textsc{\acronym{\acronym{MNIST}} 7} &\maxf{\num{0.92}} &\num{0.88} &\maxf{\num{0.92}} &\num{0.89} &$\star$ &$\star$\\
\textsc{\acronym{\acronym{MNIST}} 8} &\num{0.78} &\num{0.64} &\num{0.79} &\maxf{\num{0.84}} &$\star$ &$\star$\\
\textsc{\acronym{\acronym{MNIST}} 9} &\num{0.89} &\num{0.82} &\maxf{\num{0.90}} &\maxf{\num{0.90}} &$\star$ &$\star$\\
\textsc{\acronym{\acronym{SVHN}} 1} &\num{0.90} &\num{0.88} &\maxf{\num{0.91}} &\num{0.85} &$\star$ &$\star$\\
\textsc{\acronym{\acronym{SVHN}} 2} &\maxf{\num{0.88}} &\num{0.76} &\num{0.78} &\num{0.78} &$\star$ &$\star$\\
\textsc{\acronym{\acronym{SVHN}} 3} &\maxf{\num{0.72}} &\num{0.58} &\num{0.59} &\num{0.71} &$\star$ &$\star$\\
\textsc{\acronym{\acronym{SVHN}} 4} &\maxf{\num{0.85}} &\num{0.74} &\num{0.75} &\num{0.83} &$\star$ &$\star$\\
\textsc{\acronym{\acronym{SVHN}} 5} &\maxf{\num{0.83}} &\num{0.74} &\num{0.73} &\num{0.74} &$\star$ &$\star$\\
\textsc{\acronym{\acronym{SVHN}} 6} &\num{0.84} &\num{0.79} &\num{0.80} &\maxf{\num{0.87}} &$\star$ &$\star$\\
\textsc{\acronym{\acronym{SVHN}} 7} &\maxf{\num{0.89}} &\num{0.86} &\num{0.86} &\num{0.87} &$\star$ &$\star$\\
\textsc{\acronym{\acronym{SVHN}} 8} &\num{0.83} &\num{0.76} &\num{0.75} &\maxf{\num{0.88}} &$\star$ &$\star$\\
\textsc{\acronym{\acronym{SVHN}} 9} &\num{0.85} &\num{0.79} &\num{0.80} &\maxf{\num{0.87}} &$\star$ &$\star$\\
\textsc{Shuttle} &\num{0.99} &\maxf{\num{1.00}} &\num{0.91} &\num{0.55} &$\star$ &$\star$\\
\textsc{Satellite} &\maxf{\num{0.79}} &\num{0.70} &\num{0.62} &\num{0.57} &\num{0.78} &\num{0.74}\\
\textsc{Pima} &\maxf{\num{0.68}} &\maxf{\num{0.68}} &\num{0.62} &\num{0.59} &\num{0.67} &\num{0.65}\\
\textsc{Breastw} &\maxf{\num{0.99}} &\maxf{\num{0.99}} &\num{0.81} &\num{0.45} &\maxf{\num{0.99}} &\maxf{\num{0.99}}\\
\textsc{Arrythmia} &\num{0.79} &\maxf{\num{0.80}} &\num{0.71} &\num{0.68} &\num{0.74} &\num{0.79}\\
\textsc{Ionosphere} &\num{0.92} &\num{0.85} &\num{0.66} &\num{0.89} &\num{0.81} &\maxf{\num{0.93}}\\
\textsc{Biomed} &\num{0.87} &\num{0.83} &\num{0.76} &\num{0.69} &\maxf{\num{0.88}} &\maxf{\num{0.88}}\\
		\addlinespace
		\cmidrule(lr){2-2}\cmidrule(lr){3-3}\cmidrule(lr){4-4}\cmidrule(lr){5-5}\cmidrule(lr){6-6}\cmidrule(lr){7-7}
		\addlinespace
		\textsc{Average rank} & \maxf{\num{1.85}} & \num{3.48} &\num{2.90} &\num{3.06} &\num{4.93} &\num{4.77} \\
		\addlinespace
		\bottomrule 
		\addlinespace 
		\multicolumn{5}{l}{$\diamond$ Out of memory}\\
		\multicolumn{5}{l}{$\star$ Execution time takes more than two days}			
	\end{tabularx}
	\label{tbl:AUC:batch}	
\end{table*}

\begin{table*}
	\centering
	\newcolumntype{H}{>{\hsize=7\hsize}X}
	\newcolumntype{Y}{S[table-format=2]}
	\newcolumntype{Z}{S[table-format=5]}
	\caption{Batch anomaly detection runtimes $[t]=\si{\second}$}
	\small
	\begin{tabularx}{.95\linewidth}{HYYZZYY}
	\toprule
		& \multicolumn{1}{c}{\textsc{\EXPoSE}}  & \multicolumn{1}{c}{\textsc{\emph{i}Forest}} & \multicolumn{1}{c}{\acronym{\acronym{OC-SVM}}} & \multicolumn{1}{c}{\acronym{\acronym{LOF}}} & \multicolumn{1}{c}{\acronym{\acronym{KDE}}}  & \multicolumn{1}{c}{\textsc{FastAbod}}  \\ 
		\cmidrule(lr){2-2}\cmidrule(lr){3-3}\cmidrule(lr){4-4}\cmidrule(lr){5-5}\cmidrule(lr){6-6}\cmidrule(lr){7-7}
		\addlinespace
\textsc{\textsc{KddCup}} &44 &70 &22213 &$\diamond$ &$\star$ &$\star$\\
\textsc{ForestCover} &29 &24 &25901 &47 &$\star$ &$\star$\\
\textsc{\acronym{\acronym{MNIST}} 1} &12 &7 &1976 &23760 &$\star$ &$\star$\\
\textsc{\acronym{\acronym{MNIST}} 2} &11 &9 &2773 &18717 &$\star$ &$\star$\\
\textsc{\acronym{\acronym{MNIST}} 3} &11 &8 &1991 &20109 &$\star$ &$\star$\\
\textsc{\acronym{\acronym{MNIST}} 4} &11 &8 &1159 &17412 &$\star$ &$\star$\\
\textsc{\acronym{\acronym{MNIST}} 5} &10 &8 &1892 &15324 &$\star$ &$\star$\\
\textsc{\acronym{\acronym{MNIST}} 6} &10 &8 &3091 &18208 &$\star$ &$\star$\\
\textsc{\acronym{\acronym{MNIST}} 7} &11 &8 &2727 &20153 &$\star$ &$\star$\\
\textsc{\acronym{\acronym{MNIST}} 8} &11 &8 &1607 &18217 &$\star$ &$\star$\\
\textsc{\acronym{\acronym{MNIST}} 9} &10 &7 &2383 &18542 &$\star$ &$\star$\\
\textsc{\acronym{\acronym{SVHN}} 1} &24 &11 &9311 &18192 &$\star$ &$\star$\\
\textsc{\acronym{\acronym{SVHN}} 2} &20 &10 &10371 &18144 &$\star$ &$\star$\\
\textsc{\acronym{\acronym{SVHN}} 3} &16 &9 &14122 &28508 &$\star$ &$\star$\\
\textsc{\acronym{\acronym{SVHN}} 4} &13 &10 &4044 &19247 &$\star$ &$\star$\\
\textsc{\acronym{\acronym{SVHN}} 5} &14 &11 &10348 &22359 &$\star$ &$\star$\\
\textsc{\acronym{\acronym{SVHN}} 6} &11 &7 &5389 &13166 &$\star$ &$\star$\\
\textsc{\acronym{\acronym{SVHN}} 7} &11 &8 &6790 &14906 &$\star$ &$\star$\\
\textsc{\acronym{\acronym{SVHN}} 8} &9 &7 &4210 &9741 &$\star$ &$\star$\\
\textsc{\acronym{\acronym{SVHN}} 9} &9 &6 &3831 &9290 &$\star$ &$\star$\\
\textsc{Shuttle} &3 &7 &38 &24 &$\star$ &$\star$\\
\textsc{Satellite} &0 &3 &3 &4 &1 &55\\
\textsc{Pima} &1 &2 &0 &0 &0 &9\\
\textsc{Breastw} &1 &3 &0 &0 &0 &4\\
\textsc{Arrythmia} &1 &1 &0 &0 &0 &4\\
\textsc{Ionosphere} &1 &3 &0 &0 &0 &0\\
\textsc{Biomed} &0 &2 &0 &0 &0 &0\\
		\addlinespace
		\bottomrule 
		\addlinespace 
		\multicolumn{5}{l}{$\diamond$ Out of memory}\\
		\multicolumn{5}{l}{$\star$ Execution time takes more than two days}
	\end{tabularx}
	\label{tbl:time:batch}	
\end{table*}

\begin{figure}[b]
\begin{center}
	\includegraphics{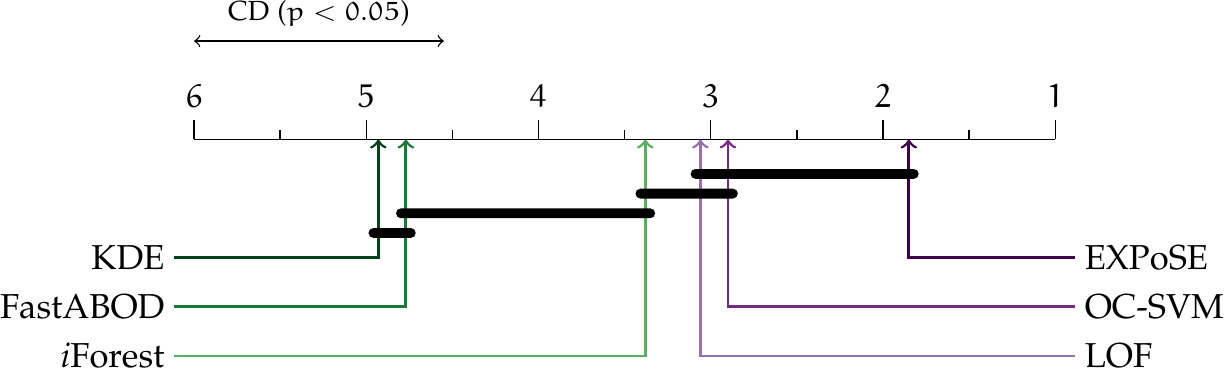}
	\caption{Critical difference diagram of the batch anomaly detection performance. Algorithms which are not significantly different (at $p < 0.05$) are connected with a bar.}
	\label{fig:critical_difference_batch}
\end{center}
\end{figure}

With the \acronym{AUC} values we can perform the Friedman and post-hoc Nemenyi tests. The Friedman test confirms a statistical significant difference between the performances of the individual algorithms at a $p$-value of $0.05$. From the critical difference diagram in \cref{fig:critical_difference_batch} we observe that \EXPoSE performs significant better
than \emph{i}Forest, FastAbod and \acronym{KDE}. 
While no significant difference in terms of anomaly detection between \EXPoSE, \acronym{OC-SVM} and \acronym{LOF} can be confirmed, \EXPoSE is several orders of magnitude faster on large-scale, high-dimensional datasets.

\clearpage
\subsection{Streaming Anomaly Detection}

In this set of experiments we compare the streaming variants of \EXPoSE against \acronym{HSTa}, \acronym{STORM} and \acronym{COD}. All of these algorithms are \emph{blind} methods as they adapt their model at regular intervals without knowing if a concept drift occurred or not. They can be combined with a concept drift detector to make the adaptation \emph{informed} \autocite{gama2010knowledge}.

The evaluation of streaming algorithms is not as straightforward as the rating of batch learning techniques. There are two accepted techniques proposed in literature.

\begin{itemize}
\item Using a dedicated subset of the data (\emph{holdout}) and evaluate the algorithm at regular time intervals. The holdout set must reflect the respective stream properties and therefore has to evolve with the stream in case of a concept drift. 
\item Making a prediction as the instance becomes available (\emph{prequential})\footnote{Prequential originates from predictive and sequential \autocite{dawid1984present}.}. A performance metric can then be applied based on the prediction and the actual label of the instance. Since predictions are made on the stream directly there are no special actions which have to be taken in case of concept drift.
\end{itemize}

If possible, the holdout method is preferable since it is an unbiased risk estimator and we can use a balanced test set with the same number of normal instances and anomalies. This is a disadvantage of the prequential method since, by definition, the data stream contains only a few anomalies. This is problematic since \acronym{STORM} and \acronym{COD} assign hard class labels and, in contrast to \acronym{AUC}, the classification \emph{accuracy} is highly sensitive to unbalanced data. We will therefore use the \emph{balanced accuracy} defined as
\begin{align*}
\frac{0.5\cdot\text{true positives}}{\text{true positives}+\text{false negatives}} + \frac{0.5\cdot\text{true negatives}}{\text{true negatives}+\text{false positives}},
\end{align*}
which compensates the unequal class distribution.

\subsubsection{Datasets}

\begin{figure}[b]
	\centering
	\includegraphics{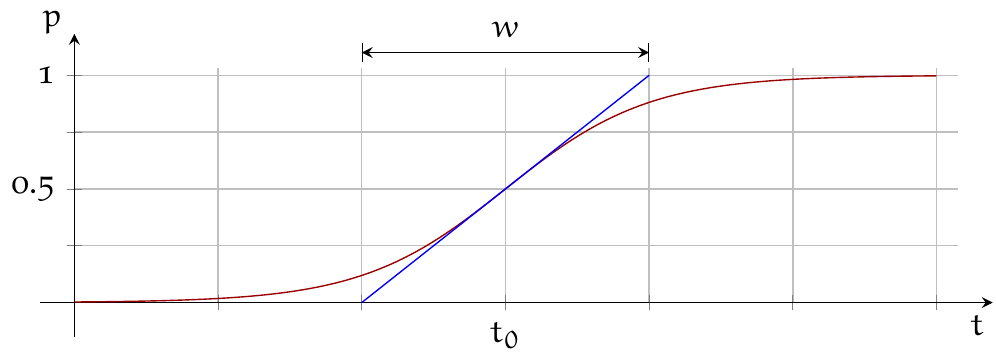}
    \caption{The sigmoid function used to introduce a smooth drift from one concept to another. We use a Bernoulli distribution, where $p$ is the probability to sample an instance from concept 1 and $(1-p)$ is the probability to sample from concept 2. The drift occurs at $t_0$ and $w$ defines the duration during which both concepts are valid.}    
    \label{fig:stream_merge}
\end{figure}

There exist only a few non-synthetic datasets for anomaly detection with concept drift. Most of them are based on multi-class datasets, where each class represents a single concept. For example we use the \acronym{SVHN} dataset and stream 9000 randomly sampled instances of the digits 1 to 9 in sequence, such that the 1000 instances of digit 1 appear first, then 1000 instances of digit 2 until digit 9 (see \cref{fig:model_adaptation}).
Every 25 time steps we calculate the accuracy using the holdout method for a dedicated random test set which contains 500 instances of the normal class and 500 instances of anomalies. Here, the normal class is the digit which is streamed at time step $t$ and anomalies are all other classes. Likewise we proceed with the Satellite and Shuttle datasets. 

Similar, \citet{ho2005martingale} proposed the \emph{three digit data stream} (\acronym{TDDS}) which contains four different concepts. Each concept consists of three digits of the \acronym{USPS} handwritten digits dataset as described\footnote{See \autocite{{ho2005martingale}} for a detailed description of the \acronym{TDDS} dataset.}.
After all instances of concept 1 are processed, the stream switches to the second concept and so on until concept 4. We randomly induce 1\% anomalies to each concept and use the prequential method for evaluation to calculate the balanced accuracy.

%\begin{table}[tb]
%\begin{center}
%	\caption{Properties of the three digit data stream (\acronym{TDDS})}
%	\small
%	\begin{tabular}{ccc} 
%		\toprule 
%		Concept & digits & concept change point\\ 
%		1 & $0,1,2$ & $1831$ \\
%		2 & $0,3,4$ & $3738$ \\
%		3 & $1,5,6$ & $5461$ \\
%		4 & $7,8,9$ & - \\
%		\midrule
%		\bottomrule 
%	\end{tabular}
%	\label{tbl:tdds_dataset}
%\end{center}
%\end{table}

All datasets presented so far contain one or more \emph{sudden} (abrupt) concept drifts. \Citet{bifet2009new} proposed a methodology to introduce a \emph{smooth} (incremental) drift between two concepts. The instances of the concepts from two classes under consideration are sampled according to a Bernoulli distribution where the class probability smoothly changes from one class to the other according to a sigmoid function (\cref{fig:stream_merge}). The concept drift occurs at $t_0$ and $w$ is the length of the drift interval. During this interval the instances of both concepts belong to the class of normal data. We apply this methodology to \acronym{USPS} and create the \emph{smooth digit drift} (\acronym{SDD}) dataset. We start with digit 1 and then smoothly change to digit 2 at $t_0 = 500$ using $w = 100$. The next drift to digit 3 occurs at $t_0 = 1000$ and we repeat this until digit 9. As before, we randomly add 1\% anomalies to each concept and use the prequential method for evaluation. We summarized the dataset characteristics in \cref{tbl:streaming_dataset}.

\begin{table}
\begin{center}
	\caption{Streaming dataset properties}
	\small
	%\begin{tabu} to .9\textwidth {X[2,l]X[c]X[c]X[c]}
	\begin{tabular}{lccc} 
		\toprule
		\addlinespace 
		 & \textsc{\#concepts} & \textsc{concept drift type} & \textsc{evaluation method}\\ 
		 \cmidrule(lr){2-2}\cmidrule(lr){3-3}\cmidrule(lr){4-4}
		 \addlinespace
		\acronym{SVHN} & 9 & sudden & holdout\\
		\textsc{Satellite} & 3 & sudden & holdout\\
		\textsc{Shuttle} & 2 & sudden & holdout\\
		\acronym{TDDS} & 4 & sudden & prequential\\
		\textsc{SDD} & 9 & smooth & prequential\\
		\addlinespace
		\bottomrule 
	\end{tabular}
	\label{tbl:streaming_dataset}
\end{center}
\end{table}

\subsubsection{Evaluation}

In the following we will denote \EXPoSE with a sliding window (\cref{sec:windowing}) and \EXPoSE with gradual forgetting (\cref{sec:gradualfrogetting}) by {$w$-\EXPoSE} and {$\gamma$-\EXPoSE}, respectively.

A sliding window of length 100 demonstrated to obey an appropriate trade off between drift adaptation and model accuracy. We therefore use this length for all algorithms and all datasets except $\gamma$-\EXPoSE. A change of the window length affects $w$-\EXPoSE, \acronym{COD}, \acronym{STORM} and \acronym{HSTa} in the same way. This is not unexpected as the window size determines the number of instances available to the algorithm.
The first 100 instances of each stream are used to configure algorithm parameters via cross-validation using pattern search. 

%\tikzexternaldisable
\begin{figure}
	\begin{center}
	\includegraphics{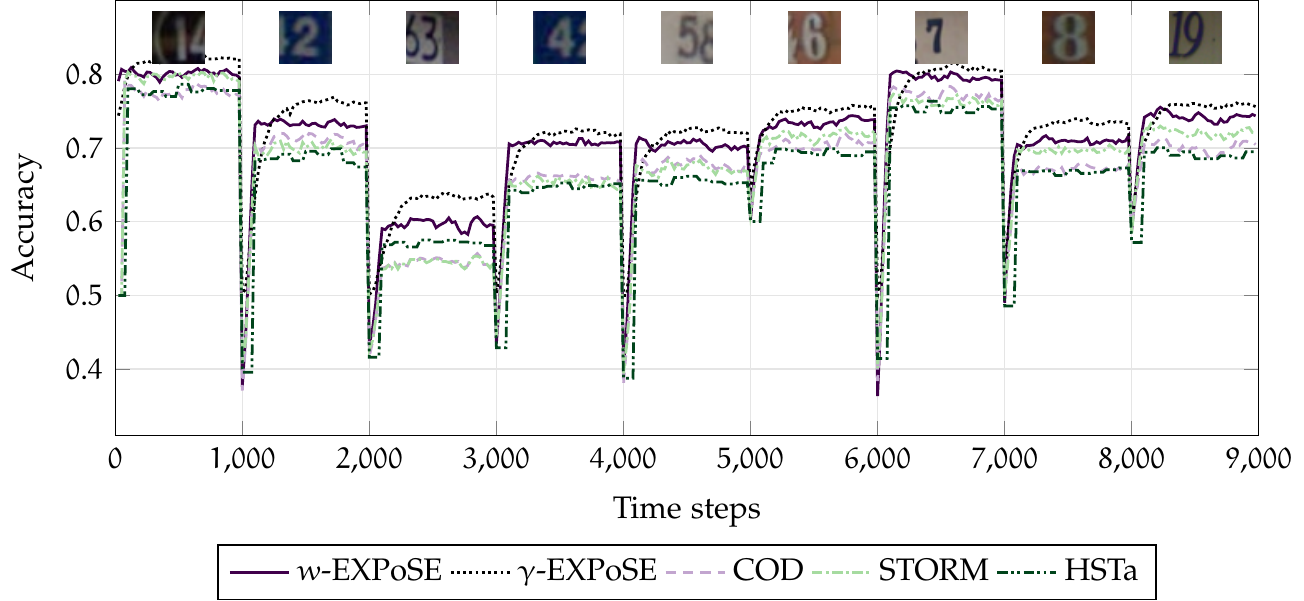}
	\end{center}
	\caption{The streaming \acronym{SVHN} experiment. A comparison of prediction accuracy under concept drift averaged over 5 repetitions.}
	\label{fig:model_adaptation}
\end{figure}
%\tikzexternalenable

\begin{table}
\begin{center}
	\caption{Streaming anomaly detection performance [accuracy]}
	\small
	\begin{tabu} to .9\textwidth {X[2,l]X[c]X[c]X[c]X[c]X[c]X[c]}
		\toprule 
		\addlinespace
		& \multicolumn{1}{c}{$w$-\EXPoSE} & \multicolumn{1}{c}{$\gamma$-\EXPoSE}   & \multicolumn{1}{c}{\acronym{STORM}} & \multicolumn{1}{c}{\acronym{COD}} & \multicolumn{1}{c}{\acronym{HSTa}} \\ 
		\cmidrule(lr){2-2}\cmidrule(lr){3-3}\cmidrule(lr){4-4}\cmidrule(lr){5-5}\cmidrule(lr){6-6}
		\addlinespace
		\textsc{Shuttle} & \num{0.88} & \num{0.88} & \num{0.75} & \num{0.74} & \maxf{\num{0.89}}\\
		\textsc{Satellite} & \maxf{\num{0.89}} & \num{0.88} & \num{0.78} & \num{0.79} & \num{0.88}\\
		\acronym{SVHN} & \num{0.71} & \maxf{\num{0.73}} & \num{0.68} & \num{0.68} & \num{0.66}\\
		\acronym{TDDS} & \num{0.71} & \maxf{\num{0.71}} & \num{0.67} & \num{0.64} & \num{0.67}\\
		\acronym{SDD} & \num{0.83} & \maxf{\num{0.85}} & \num{0.79} & \num{0.76} & \num{0.77}\\
		\addlinespace
		\cmidrule(lr){2-2}\cmidrule(lr){3-3}\cmidrule(lr){4-4}\cmidrule(lr){5-5}\cmidrule(lr){6-6}
		\addlinespace
		\textsc{Average rank} & \num{1.80} & \maxf{\num{1.70}} &\num{3.80} &\num{4.50} &\num{3.20}  \\	
		\addlinespace	
		\bottomrule 	  
	\end{tabu}
	\label{tbl:cc:streaming}
\end{center}
\end{table}

%\tikzexternaldisable
\begin{figure}[tbp]
	\centering
	\includegraphics{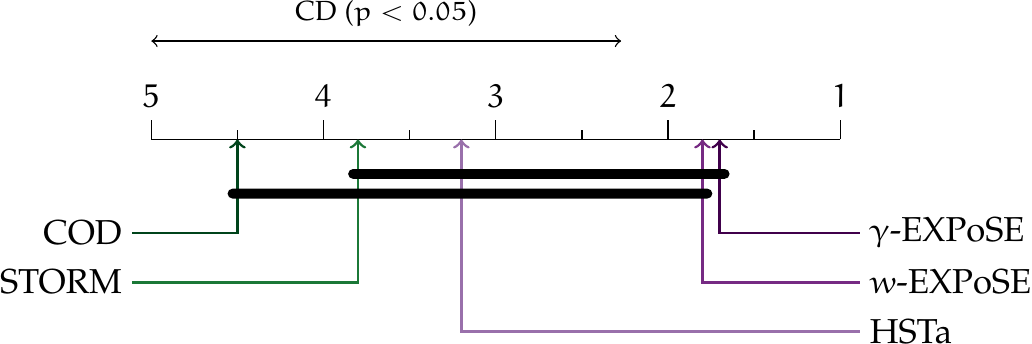}
	\caption[Nemenyi test stream]{Critical difference diagram of the stream anomaly detection performance. Algorithms which are not significantly different (at $p < 0.05$) are connected with a bar.}
	\label{fig:critical_difference_stream}
\end{figure}
%\tikzexternalenable

A detailed illustration of the \acronym{SVHN} experiment is shown in \cref{fig:model_adaptation}.
The predictive performance of all algorithms is relatively similar. 
It can be observed that, as the stream changes from one digit to another, the accuracy suddenly drops which indicates that the current model is not valid anymore. After a short period of time, the model adapts and the accuracy recovers. \EXPoSE performs on average better than \acronym{COD}, \acronym{STORM} and \acronym{HSTa}. A possible interpretation of this result is the sound foundation in probability theory of our approach.
The suboptimal performance of \acronym{HSTa} indicates the random binary trees constructed by \acronym{HSTa} are not sufficiently fine-grained for this high-dimensional datasets. This interpretation is supported by the experiments with the low-dimensional Shuttle and Satellite data, where \acronym{HSTa} performs better.

The average over all accuracies of the individual experiments can be found in \cref{tbl:cc:streaming}. The only statistical significance (at $p<0.05$) is observed between $\gamma$-\EXPoSE and \acronym{COD}. We could not confirm a significant difference between the other algorithms as illustrated in the critical difference diagram (\cref{fig:critical_difference_stream}).

Although these results are promising we recommend to combine the techniques presented here with a concept drift detection technique to make \emph{informed} model updates \autocite{gama2010knowledge}.

\section{Conclusion}

We proposed a new algorithm, \emph{\EXPoSE}, to perform anomaly detection on very large-scale datasets and streams with concept drift. Although anomaly detection is a problem of central importance in many applications, only a few algorithms are scalable to the vast amount of data we are often confronted with.

The \EXPoSE anomaly detection classifier calculates a score (the likelihood of a query point belonging to the class of normal data) using the inner product between a feature map and the kernel embedding of probability measures. The kernel embedding technique provides an efficient way to work with probability measures without the necessity to make assumptions about the underlying distributions.

Despite its simplicity \EXPoSE obeys a \emph{linear} computational complexity for learning and can make predictions in \emph{constant} time while it requires only  \emph{constant} memory. 
When applied incrementally or online, a model update can also be performed in \emph{constant} time. We demonstrated that \EXPoSE can be used as an efficient anomaly detection algorithm with the same predictive performance as the best state of the art methods while being significant faster than techniques with the same discriminant power.


\clearpage
\begin{thebibliography}{68}
\providecommand{\natexlab}[1]{#1}
\providecommand{\url}[1]{{#1}}
\providecommand{\urlprefix}{URL }
\expandafter\ifx\csname urlstyle\endcsname\relax
  \providecommand{\doi}[1]{DOI~\discretionary{}{}{}#1}\else
  \providecommand{\doi}{DOI~\discretionary{}{}{}\begingroup
  \urlstyle{rm}\Url}\fi
\providecommand{\eprint}[2][]{\url{#2}}

\bibitem[{Aleskerov et~al(1997)Aleskerov, Freisleben, and Rao}]{Aleskerov1997}
Aleskerov E, Freisleben B, Rao B (1997) {Cardwatch: A neural network based
  database mining system for credit card fraud detection}. In: Computational
  Intelligence for Financial Engineering, \doi{10.1109/cifer.1997.618940}

\bibitem[{Aliprantis and Border(2006)}]{aliprantis2006infinite}
Aliprantis CD, Border K (2006) {Infinite dimensional analysis: a hitchhiker's
  guide}. Springer Science {\&} Business Media, \doi{10.1007/3-540-29587-9}

\bibitem[{Angiulli and Fassetti(2007)}]{angiulli2007detecting}
Angiulli F, Fassetti F (2007) {Detecting distance-based outliers in streams of
  data}. In: Proceedings of the sixteenth ACM conference on Conference on
  information and knowledge management, ACM, pp 811--820,
  \doi{10.1145/1321440.1321552}

\bibitem[{Angiulli and Fassetti(2010)}]{angiulli2010distance}
Angiulli F, Fassetti F (2010) {Distance-based outlier queries in data streams:
  the novel task and algorithms}. Data Mining and Knowledge Discovery
  20(2):290--324, \doi{10.1007/s10618-009-0159-9}

\bibitem[{Angiulli and Pizzuti(2002)}]{angiulli2002fast}
Angiulli F, Pizzuti C (2002) {Fast outlier detection in high dimensional
  spaces}. In: PKDD, Springer, vol~2, pp 15--26, \doi{10.1007/3-540-45681-3_2}

\bibitem[{Bifet et~al(2009)Bifet, Holmes, Pfahringer, Kirkby, and
  Gavald{\`{a}}}]{bifet2009new}
Bifet A, Holmes G, Pfahringer B, Kirkby R, Gavald{\`{a}} R (2009) {New ensemble
  methods for evolving data streams}. In: Proceedings of the 15th ACM SIGKDD
  international conference on Knowledge discovery and data mining, ACM, pp
  139--148, \doi{10.1145/1557019.1557041}

\bibitem[{Breunig et~al(2000)Breunig, Kriegel, Ng, and Sander}]{Breunig2000}
Breunig MM, Kriegel HP, Ng RT, Sander J (2000) {LOF: identifying density-based
  local outliers}. In: ACM Sigmod Record, ACM, vol~29, pp 93--104,
  \doi{10.1145/335191.335388}

\bibitem[{Chandola et~al(2009)Chandola, Banerjee, and Kumar}]{Chandola2009}
Chandola V, Banerjee A, Kumar V (2009) {Anomaly detection: A survey}. ACM
  Computing Surveys (CSUR) 41(3):1--58

\bibitem[{Chang and Lin(2011)}]{Chang2011}
Chang CC, Lin CJ (2011) {LIBSVM: A library for support vector machines}. ACM
  Transactions on Intelligent Systems and Technology 2(3):1--27

\bibitem[{Dawid(1984)}]{dawid1984present}
Dawid AP (1984) {Present position and potential developments: Some personal
  views: Statistical theory: The prequential approach}. Journal of the Royal
  Statistical Society Series A (General) pp 278--292, \doi{10.2307/2981683}

\bibitem[{Dean and Ghemawat(2008)}]{dean2008mapreduce}
Dean J, Ghemawat S (2008) {MapReduce: simplified data processing on large
  clusters}. Communications of the ACM 51(1):107--113,
  \doi{10.1145/1327452.1327492}

\bibitem[{Dem{\v{s}}ar(2006)}]{demvsar2006statistical}
Dem{\v{s}}ar J (2006) {Statistical comparisons of classifiers over multiple
  data sets}. The Journal of Machine Learning Research 7:1--30

\bibitem[{Domingos and Hulten(2000)}]{domingos2000mining}
Domingos P, Hulten G (2000) {Mining high-speed data streams}. In: Proceedings
  of the sixth ACM SIGKDD international conference on Knowledge discovery and
  data mining, ACM, pp 71--80, \doi{10.1145/347090.347107}

\bibitem[{Domingos and Hulten(2001)}]{domingos2001catching}
Domingos P, Hulten G (2001) {Catching up with the Data: Research Issues in
  Mining Data Streams}. In: DMKD

\bibitem[{Faria et~al(2013)Faria, Gama, and Carvalho}]{faria2013novelty}
Faria ER, Gama J, Carvalho AC (2013) {Novelty detection algorithm for data
  streams multi-class problems}. In: Proceedings of the 28th Annual ACM
  Symposium on Applied Computing, ACM, pp 795--800,
  \doi{10.1145/2480362.2480515}

\bibitem[{Friedman(1937)}]{friedman1937use}
Friedman M (1937) {The use of ranks to avoid the assumption of normality
  implicit in the analysis of variance}. Journal of the American Statistical
  Association 32(200):675--701, \doi{10.1080/01621459.1937.10503522}

\bibitem[{Gama(2010)}]{gama2010knowledge}
Gama J (2010) {Knowledge discovery from data streams}. CRC Press,
  \doi{10.1201/ebk1439826119-c1}

\bibitem[{Gama et~al(2014)Gama, {\v{Z}}liobaite, Bifet, Pechenizkiy, and
  Bouchachia}]{gama2014survey}
Gama J, {\v{Z}}liobaite I, Bifet A, Pechenizkiy M, Bouchachia A (2014) {A
  survey on concept drift adaptation}. ACM Computing Surveys (CSUR) 46(4):44,
  \doi{10.1145/2523813}

\bibitem[{Goodman and O'Rourke(2004)}]{Goodman2004}
Goodman JE, O'Rourke J (2004) {Handbook of discrete and computational
  geometry}, 2nd edn. CRC press

\bibitem[{Gretton and Desobry(2003)}]{gretton2003line}
Gretton A, Desobry F (2003) {On-line one-class support vector machines. an
  application to signal segmentation}. In: Acoustics, Speech, and Signal
  Processing, 2003. Proceedings.(ICASSP'03). 2003 IEEE International Conference
  on, IEEE, vol~2, pp 2--709, \doi{10.1109/icassp.2003.1202465}

\bibitem[{Gretton et~al(2012)Gretton, Borgwardt, Rasch, Sch{\"{o}}lkopf, and
  Smola}]{Gretton2012}
Gretton A, Borgwardt KM, Rasch MJ, Sch{\"{o}}lkopf B, Smola A (2012) {A kernel
  two-sample test}. The Journal of Machine Learning Research 13(1):723--773

\bibitem[{Gupta et~al(2014)Gupta, Gao, Aggarwal, and Han}]{Gupta2014}
Gupta M, Gao J, Aggarwal CC, Han J (2014) {Outlier Detection for Temporal Data:
  A Survey}. Knowledge and Data Engineering, IEEE Transactions on
  26(9):2250--2267, \doi{10.1109/tkde.2013.184}

\bibitem[{Ho(2005)}]{ho2005martingale}
Ho SS (2005) {A martingale framework for concept change detection in
  time-varying data streams}. In: Proceedings of the 22nd international
  conference on Machine learning, ACM, pp 321--327,
  \doi{10.1145/1102351.1102392}

\bibitem[{Hodge and Austin(2004)}]{Hodge2004}
Hodge VJ, Austin J (2004) {A survey of outlier detection methodologies}.
  Artificial Intelligence Review 22(2):85--126,
  \doi{10.1023/b:aire.0000045502.10941.a9}

\bibitem[{Iman and Davenport(1980)}]{iman1980approximations}
Iman RL, Davenport JM (1980) {Approximations of the critical region of the
  Friedman statistic}. Communications in Statistics-Theory and Methods
  9(6):571--595

\bibitem[{Kallenberg(2006)}]{kallenberg2006foundations}
Kallenberg O (2006) {Foundations of modern probability}. Springer Science {\&}
  Business Media, \doi{10.1007/b98838}

\bibitem[{Kar and Karnick(2012)}]{kar2012random}
Kar P, Karnick H (2012) {Random feature maps for dot product kernels}.
  International Conference on Artificial Intelligence and Statistics pp
  583--591

\bibitem[{Knorr and Ng(1998)}]{knorr1998algorithms}
Knorr EM, Ng RT (1998) {Algorithms for mining distance-based outliers in large
  datasets}. In: Proceedings of the International Conference on Very Large Data
  Bases, Citeseer, pp 392--403

\bibitem[{Knorr et~al(2000)Knorr, Ng, and Tucakov}]{knorr2000distance}
Knorr EM, Ng RT, Tucakov V (2000) {Distance-based outliers: algorithms and
  applications}. The VLDB Journal—The International Journal on Very Large
  Data Bases 8(3-4):237--253, \doi{10.1007/s007780050006}

\bibitem[{Kontaki et~al(2011)Kontaki, Gounaris, Papadopoulos, Tsichlas, and
  Manolopoulos}]{kontaki2011continuous}
Kontaki M, Gounaris A, Papadopoulos AN, Tsichlas K, Manolopoulos Y (2011)
  {Continuous monitoring of distance-based outliers over data streams}. In:
  Data Engineering (ICDE), 2011 IEEE 27th International Conference on, IEEE, pp
  135--146, \doi{10.1109/icde.2011.5767923}

\bibitem[{Kriegel and Zimek(2008)}]{kriegel2008angle}
Kriegel HP, Zimek A (2008) {Angle-based outlier detection in high-dimensional
  data}. In: Proceedings of the 14th ACM SIGKDD international conference on
  Knowledge discovery and data mining, ACM, pp 444--452,
  \doi{10.1145/1401890.1401946}

\bibitem[{Kumar(2005)}]{Kumar2005}
Kumar V (2005) {Parallel and distributed computing for cybersecurity}. IEEE
  Distributed Systems Online 6(10):1, \doi{10.1109/mdso.2005.53}

\bibitem[{Lazarescu et~al(2004)Lazarescu, Venkatesh, and
  Bui}]{lazarescu2004using}
Lazarescu MM, Venkatesh S, Bui HH (2004) {Using multiple windows to track
  concept drift}. Intelligent data analysis 8(1):29--59

\bibitem[{Le et~al(2013)Le, Sarl{\'{o}}s, and Smola}]{Le2013}
Le Q, Sarl{\'{o}}s T, Smola AJ (2013) {Fastfood: approximating kernel
  expansions in loglinear time}. In: Proceedings of the international
  conference on machine learning,
  

\bibitem[{Li et~al(2010)Li, Ionescu, and Sminchisescu}]{li2010random}
Li F, Ionescu C, Sminchisescu C (2010) {Random Fourier approximations for
  skewed multiplicative histogram kernels}. In: Pattern Recognition, Springer,
  pp 262--271, \doi{10.1007/978-3-642-15986-2_27}

\bibitem[{Lichman(2013)}]{Lichman:2013}
Lichman M (2013) {UCI Machine Learning Repository}.
  

\bibitem[{Liu et~al(2012)Liu, Ting, and Zhou}]{liu2012isolation}
Liu FT, Ting KM, Zhou ZH (2012) {Isolation-based anomaly detection}. ACM
  Transactions on Knowledge Discovery from Data 6(1):3,
  \doi{10.1145/2133360.2133363}

\bibitem[{Nemenyi(1963)}]{nemenyi1962distribution}
Nemenyi P (1963) {Distribution-free multiple comparisons}. PhD thesis,
  Princeton University.

\bibitem[{Netzer et~al(2011)Netzer, Wang, Coates, Bissacco, Wu, and
  Ng}]{Netzer2011}
Netzer Y, Wang T, Coates A, Bissacco A, Wu B, Ng A (2011) {Reading digits in
  natural images with unsupervised feature learning}. In: NIPS workshop on deep
  learning and unsupervised feature learning, vol 2011, p~4

\bibitem[{Ott et~al(2014)Ott, Pang, Ramos, and Chawla}]{ott2014integrated}
Ott L, Pang L, Ramos FT, Chawla S (2014) {On Integrated Clustering and Outlier
  Detection}. In: Advances in Neural Information Processing Systems, pp
  1359--1367

\bibitem[{Pokrajac(2007)}]{Pokrajac2007}
Pokrajac D (2007) {Incremental local outlier detection for data streams}. IEEE
  Symposium on Computational Intelligence and Data Mining pp 504--515,
  \doi{10.1109/cidm.2007.368917}

\bibitem[{Rahimi and Recht(2007)}]{Rahimi2007}
Rahimi A, Recht B (2007) {Random features for large-scale kernel machines}. In:
  Advances in neural information processing systems, pp 1177--1184,
  

\bibitem[{Rahimi and Recht(2008)}]{Rahimi2008}
Rahimi A, Recht B (2008) {Weighted sums of random kitchen sinks: Replacing
  minimization with randomization in learning}. In: Advances in neural
  information processing systems, pp 1313--1320,

\bibitem[{Ramaswamy et~al(2000)Ramaswamy, Rastogi, and
  Shim}]{ramaswamy2000efficient}
Ramaswamy S, Rastogi R, Shim K (2000) {Efficient algorithms for mining outliers
  from large data sets}. In: ACM SIGMOD Record, ACM, vol~29, pp 427--438,
  \doi{10.1145/335191.335437}

\bibitem[{Sadik and Gruenwald(2014)}]{sadik2014research}
Sadik S, Gruenwald L (2014) {Research issues in outlier detection for data
  streams}. ACM SIGKDD Explorations Newsletter 15(1):33--40,
  \doi{10.1145/2594473.2594479}

\bibitem[{Schneider(2016)}]{Schneider2016}
Schneider M (2016) {Probability Inequalities for Kernel Embeddings in Sampling
  without Replacement}. In: Proceedings of the Nineteenth International
  Conference on Artificial Intelligence and Statistics

\bibitem[{Schneider et~al(2015)Schneider, Ertel, and Palm}]{Schneider2015a}
Schneider M, Ertel W, Palm G (2015) {Expected Similarity Estimation for Large
  Scale Anomaly Detection}. In: International Joint Conference on Neural
  Networks, IEEE, pp 1--8, \doi{10.1109/ijcnn.2015.7280331}

\bibitem[{Sch{\"{o}}lkopf et~al(2001)Sch{\"{o}}lkopf, Platt, Shawe-Taylor,
  Smola, and Williamson}]{Scholkopf2001}
Sch{\"{o}}lkopf B, Platt JC, Shawe-Taylor J, Smola AJ, Williamson RC (2001)
  {Estimating the support of a high-dimensional distribution}. Neural
  computation 13(7):1443--1471, \doi{10.1162/089976601750264965}

\bibitem[{Sejdinovic et~al(2013)Sejdinovic, Sriperumbudur, Gretton, and
  Fukumizu}]{sejdinovic2013equivalence}
Sejdinovic D, Sriperumbudur B, Gretton A, Fukumizu K (2013) {Equivalence of
  distance-based and RKHS-based statistics in hypothesis testing}. The Annals
  of Statistics 41(5):2263--2291, \doi{10.1214/13-aos1140}

\bibitem[{Smola et~al(2007)Smola, Gretton, Song, and
  Sch{\"{o}}lkopf}]{Smola2007}
Smola AJ, Gretton A, Song L, Sch{\"{o}}lkopf B (2007) {A Hilbert space
  embedding for distributions}. In: Algorithmic Learning Theory, Springer, pp
  13--31, \doi{10.1007/978-3-540-75488-6_5}

\bibitem[{Spence et~al(2001)Spence, Parra, and Sajda}]{Spence2001}
Spence C, Parra L, Sajda P (2001) {Detection, synthesis and compression in
  mammographic image analysis with a hierarchical image probability model}. In:
  Mathematical Methods in Biomedical Image Analysis, 2001. MMBIA 2001. IEEE
  Workshop on, IEEE, pp 3--10, \doi{10.1109/mmbia.2001.991693}

\bibitem[{Spinosa et~al(2007)Spinosa, {de Leon F de Carvalho}, and
  Gama}]{spinosa2007olindda}
Spinosa EJ, {de Leon F de Carvalho} AP, Gama J (2007) {OLINDDA: a cluster-based
  approach for detecting novelty and concept drift in data streams}. In:
  Proceedings of the 2007 ACM symposium on Applied computing, ACM, pp 448--452

\bibitem[{Steinwart(2003)}]{Steinwart2003}
Steinwart I (2003) {Sparseness of support vector machines}. The Journal of
  Machine Learning Research 4:1071--1105

\bibitem[{Steinwart and Christmann(2008)}]{steinwart2008support}
Steinwart I, Christmann A (2008) {Support vector machines}. Springer

\bibitem[{Tan et~al(2011)Tan, Ting, and Liu}]{tan2011fast}
Tan SC, Ting KM, Liu TF (2011) {Fast anomaly detection for streaming data}. In:
  IJCAI Proceedings-International Joint Conference on Artificial Intelligence,
  Citeseer, vol~22, p 1511

\bibitem[{Tax(2001)}]{Tax2001}
Tax DMJ (2001) {One-class classification}. PhD thesis, Technische Universiteit
  Delft

\bibitem[{Tax and Duin(2004)}]{Tax2004}
Tax DMJ, Duin RPW (2004) {Support vector data description}. Machine learning
  54(1)

\bibitem[{Torczon(1997)}]{torczon1997convergence}
Torczon V (1997) {On the convergence of pattern search algorithms}. SIAM
  Journal on optimization 7(1):1--25

\bibitem[{Vedaldi and Zisserman(2012)}]{vedaldi2012efficient}
Vedaldi A, Zisserman A (2012) {Efficient Additive Kernels via Explicit Feature
  Maps}. Pattern Analysis and Machine Intelligence, IEEE Transactions on
  34(3):480--492, \doi{10.1109/cvpr.2010.5539949}

\bibitem[{Vondrick et~al(2013)Vondrick, Khosla, Malisiewicz, and
  Torralba}]{vondrick2013hoggles}
Vondrick C, Khosla A, Malisiewicz T, Torralba A (2013) {Hoggles: Visualizing
  object detection features}. In: Computer Vision (ICCV), 2013 IEEE
  International Conference on, IEEE, pp 1--8, \doi{10.1109/iccv.2013.8}

\bibitem[{Webb(2000)}]{webb2000multiboosting}
Webb GI (2000) {Multiboosting: A technique for combining boosting and wagging}.
  Machine learning 40(2):159--196

\bibitem[{Widmer and Kubat(1996)}]{widmer1996learning}
Widmer G, Kubat M (1996) {Learning in the presence of concept drift and hidden
  contexts}. Machine learning 23(1):69--101, \doi{10.1007/bf00116900}

\bibitem[{Williams and Seeger(2001)}]{williams2001using}
Williams C, Seeger M (2001) {Using the Nystr{\"{o}}m method to speed up kernel
  machines}. In: Proceedings of the 14th Annual Conference on Neural
  Information Processing Systems, MIT Press, EPFL-CONF-161322, pp 682--688

\bibitem[{Wu et~al(2014)Wu, Zhang, Fan, Edwards, and Yu}]{wu2014rs}
Wu K, Zhang K, Fan W, Edwards A, Yu PS (2014) {RS-Forest: A Rapid Density
  Estimator for Streaming Anomaly Detection}. In: Data Mining (ICDM), 2014 IEEE
  International Conference on, IEEE, pp 600--609, \doi{10.1109/icdm.2014.45}

\bibitem[{Xiong et~al(2011)Xiong, Poczos, Schneider, Connolly, and
  VanderPlas}]{Xiong2011}
Xiong L, Poczos B, Schneider J, Connolly A, VanderPlas J (2011) {Hierarchical
  Probabilistic Models for Group Anomaly Detection}. In: AISTATS 2011

\bibitem[{Yamanishi et~al(2004)Yamanishi, Takeuchi, Williams, and
  Milne}]{yamanishi2004line}
Yamanishi K, Takeuchi JI, Williams G, Milne P (2004) {On-line unsupervised
  outlier detection using finite mixtures with discounting learning
  algorithms}. Data Mining and Knowledge Discovery 8(3):275--300,
  \doi{10.1145/347090.347160}

\bibitem[{Yu et~al(2003)Yu, Yang, and Han}]{yu2003classifying}
Yu H, Yang J, Han J (2003) {Classifying large data sets using SVMs with
  hierarchical clusters}. In: Proceedings of the ninth ACM SIGKDD international
  conference on Knowledge discovery and data mining, ACM, pp 306--315,
  \doi{10.1145/956750.956786}

\bibitem[{Zhang et~al(2011)Zhang, Sconyers, Byington, Patrick, Orchard, and
  Vachtsevanos}]{zhang2011probabilistic}
Zhang B, Sconyers C, Byington C, Patrick R, Orchard ME, Vachtsevanos G (2011)
  {A probabilistic fault detection approach: application to bearing fault
  detection}. IEEE Transactions on Industrial Electronics 58(5):2011--2018,
  \doi{10.1109/tie.2010.2058072}

\end{thebibliography}
\end{document}